\documentclass[10pt,twocolumn,letterpaper]{article}

\usepackage[pagenumbers]{cvpr} 

\usepackage{booktabs}     
\usepackage{multirow}     
\usepackage[table]{xcolor} 
\usepackage{colortbl}
\usepackage{amsmath}

\usepackage{algorithm}
\usepackage{algpseudocode}

\usepackage{lipsum}

\algnewcommand{\LineComment}[1]{\State \textit{// #1}}

\usepackage{siunitx}

\usepackage{xcolor}

\definecolor{cvprblue}{rgb}{0.21,0.49,0.74}
\usepackage[pagebackref,breaklinks,colorlinks,allcolors=cvprblue]{hyperref}

\def\paperID{}
\def\confName{CVPR}
\def\confYear{2026}

\title{FlowC2S: Flowing from Current to Succeeding Frames for Fast and Memory-Efficient Video Continuation}

\author{
Hovhannes Margaryan$^{1}$ \quad
Quentin Bammey$^{2}$ \quad
Christian Sandor $^{1}$\\[0.7em]
$^{1}$Team ARAI, Université Paris-Saclay, CNRS, LISN, France  \\ $^{2}$ LTCI, Télécom Paris, Institut Polytechnique de Paris, France
}

\begin{document}

\twocolumn[{%
\renewcommand\twocolumn[1][]{#1}%
\maketitle
\begin{center}
    \centering
    \captionsetup{type=figure}
    \includegraphics[width=1\linewidth]{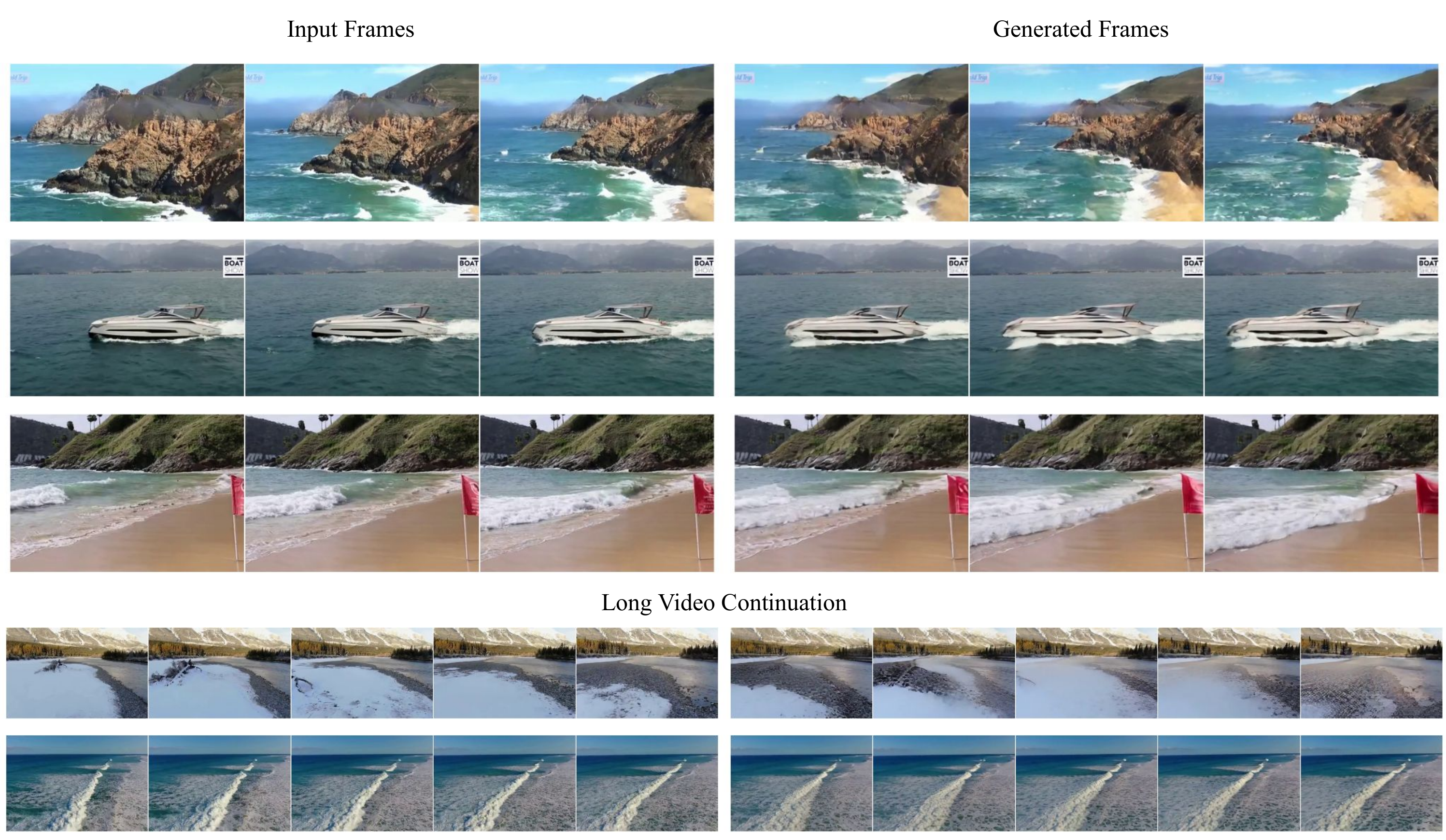}
    \captionof{figure}{FlowC2S generates video continuations starting the generation directly from the given frames. We  achieve this by training a flow matching model between current and succeeding frame distributions with inherent optimal couplings and target inversion. Our method successfully halves the model input dimensionality, generates high-quality video continuations with only five sampling steps, and generalizes to video durations beyond those seen at training. See the appendix for more examples.}
    \label{fig:fig_teaser}
\end{center}%
}]




\begin{abstract}

This paper introduces a novel methodology for generating fast and memory-efficient video continuations. Our method, dubbed FlowC2S, fine-tunes a pre-trained text-to-video flow model to learn a vector field between the current and succeeding video chunks. Two design choices are key. First, we introduce inherent optimal couplings, utilizing temporally adjacent video chunks during training as a practical proxy for true optimal couplings, resulting in straighter flows. Second, we incorporate target inversion, injecting the inverted latent of the target chunk into the input representation to strengthen correspondences and improve visual fidelity. By flowing directly from current to succeeding frames, instead of the common combination of current frames with noise to generate a video continuation, we reduce the dimensionality of the model input by a factor of two. The proposed method, fine-tuned from LTXV and Wan, surpasses the state-of-the-art scores across quantitative evaluations with FID and FVD, with as few as five neural function evaluations. The code of FlowC2S is available at \href{https://github.com/marghovo/FlowC2S}{https://github.com/marghovo/FlowC2S}.



\end{abstract}

\section{Introduction}
\label{sec:sec_0}

Recent advances in image and video generation \cite{Rombach_2022_CVPR, podell2023sdxlimprovinglatentdiffusion, esser2024scalingrectifiedflowtransformers, dai2023emuenhancingimagegeneration, xie2024sanaefficienthighresolutionimage, zheng2024opensorademocratizingefficientvideo, peng2025opensora20trainingcommerciallevel, chen2025gokuflowbasedvideo, seawead2025seaweed7bcosteffectivetrainingvideo} have dramatically improved the quality of synthetic media, largely due to diffusion- and flow-based models \cite{pmlr-v37-sohl-dickstein15, song2021denoising, NEURIPS2020_4c5bcfec, song2021scorebased, NEURIPS2021_49ad23d1,  liu2022flowstraightfastlearning, lipman2023flowmatchinggenerativemodeling, karras2024analyzingimprovingtrainingdynamics}. 
Building on this progress, researchers have turned to \textit{video continuation}: synthesizing a plausible sequence of frames given the current segment of a video. In essence, the goal of video continuation is to generate a temporally coherent and semantically consistent continuation of the input video segment. Success in video continuation would enable key applications in long-horizon video generation, world models, and immersive AR/VR environments. Indeed, AR/VR applications must circumvent the lag introduced by the use of generative models. For example, editing a future frame instead of the current one ensures that it is ready when its timestamp arrives, keeping AR/VR displays synchronized with the real world despite model latency. 



However, generating a video continuation remains difficult. Maintaining a plausible extension of a visual sequence requires balancing two sometimes-competing objectives:
temporal coherence, to ensure realistic motion and smooth dynamics, and
visual fidelity, to preserve sharp details and semantic consistency across frames. 
To tackle this, state-of-the-art methods condition on the given frames and use Gaussian noise to generate a continuation \cite{gao2024vista, hassan2024gemgeneralizableegovisionmultimodal, hacohen2024ltxvideorealtimevideolatent, yin2025causvid}. While effective, this design has key drawbacks: \textit{high memory cost} since the model must process both the conditioning frames and noise of output dimensionality, and \textit{slow inference} as multiple neural function evaluations (NFEs) are required. This is even more of a problem considering that applications such as AR/VR, which rely on video continuation as a building block, require the methods to operate on embedded devices in near-real time. 

In this work, we introduce \textbf{FlowC2S}, an alternative to the video continuation task. Our model receives a chunk of video frames as input and generates the subsequent video chunk as a plausible continuation of the given input chunk. It preserves semantic coherence with the input while propagating global contextual information. \Cref{fig:fig_teaser} summarizes our results.



\begin{figure*}[t]
        \centering
    \subfloat[Conventional input design like LTXVCondition  \cite{hacohen2024ltxvideorealtimevideolatent}.]{%
        \includegraphics[width=0.45\linewidth]{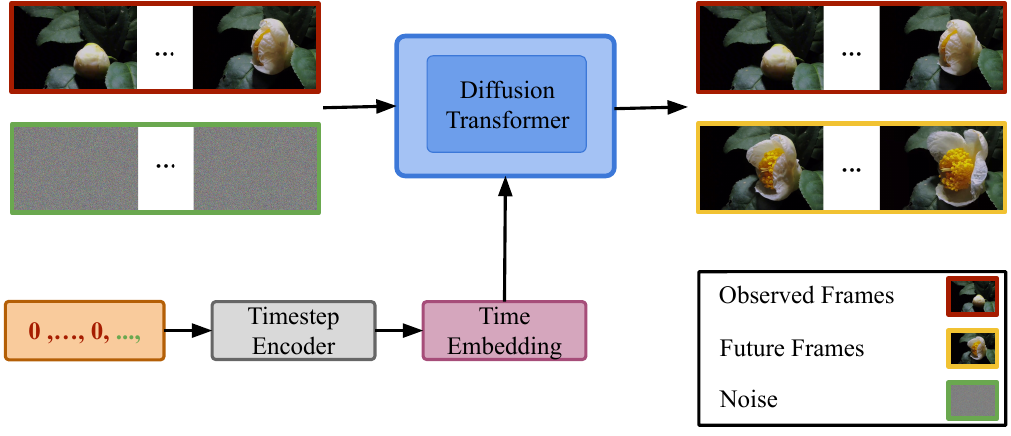}%
        \label{fig:fig_0_a}
    }\hfill
    \subfloat[Proposed input design.]{%
        \includegraphics[width=0.45\linewidth]{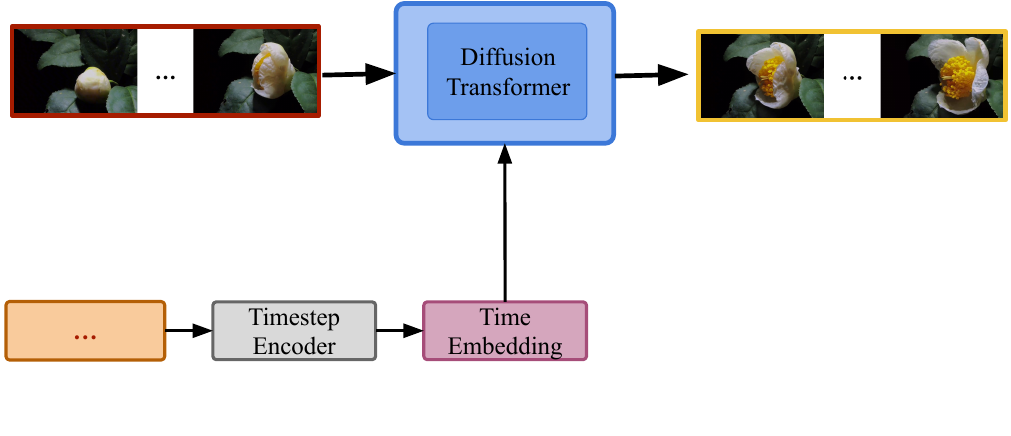}%
        \label{fig:fig_0_b}
    }
    \caption{(a) In conventional approaches, both the given input frames and additional Gaussian noise of the same size as the  chunk to be generated are provided to the model at inference. Zero timesteps are used for the conditioning frames. (b) \textbf{FlowC2S} removes the noise; the model directly learns a vector field from current to succeeding frames and thus enables sampling from solely an input video chunk. 
    }
 
    \label{fig:fig_0}
\end{figure*}

FlowC2S fine-tunes a pre-trained flow-based text-to-video model (e.g. LTXV  \cite{hacohen2024ltxvideorealtimevideolatent} or Wan  \cite{Wan2025Wanopenadvancedlargescale}) by modifying the initial distribution from the multivariate normal distribution to the distribution of current (or observed) frames. Then it learns a mapping between the input and succeeding frame distributions by flow matching  \cite{lipman2023flowmatchinggenerativemodeling, liu2022flowstraightfastlearning, albergo2023buildingnormalizingflowsstochastic}. By \textit{flowing directly from current to succeeding frames}, FlowC2S bypasses the conventional paradigm of conditioning the model on input frames and using Gaussian noise to generate a video continuation. This results in a twofold reduction of model input dimensionality. Additionally, our model is trained using inherent optimal couplings, where the input and the following video chunk from the same video serve as the source and target, respectively. This approach leads to straighter flow trajectories and hence reduces the number of neural function evaluations required at inference to outperform prior methods quantitatively. FlowC2S also uses inverted latents of the target video chunk, achieving improved visual quality. Our approach is capable of generating coherent video continuations without explicit temporal conditioning on the given input video chunk. Fig.~\ref{fig:fig_0} illustrates the key difference between the input design adopted by conventional methods and FlowC2S.


The primary contributions of this paper are:
\begin{itemize}
\item We propose a simple yet novel and efficient approach to video continuation by directly flowing from current to succeeding frames, leveraging \(2\times\) more memory-efficient input design, which yields~--~on the LTXV backbone as an example~--~approximately  \(50\%\) lower peak GPU memory relative to LTXVCondition  \cite{hacohen2024ltxvideorealtimevideolatent}.
\item Our method leverages inherent optimal couplings of input and succeeding frames as a proxy for the true optimal couplings, reducing the required neural function evaluations for state-of-the-art quantitative results (e.g., 4× fewer NFEs on LTXV backbone). In addition, we leverage target inversion for enhanced visual fidelity. 
\item Through extensive experiments on multiple datasets \cite{nan2024openvid, caesar2020nuscenesmultimodaldatasetautonomous} and backbones \cite{hacohen2024ltxvideorealtimevideolatent, Wan2025Wanopenadvancedlargescale}, we show that FlowC2S achieves competitive visual quality while substantially improving both GPU memory usage and NFEs.

\end{itemize}

\section{Related Work}
\label{sec:sec_1}





\textbf{Text-to-Video Models.} Text-to-video (T2V) diffusion  \cite{wang2023modelscopetexttovideotechnicalreport, wang2023laviehighqualityvideogeneration, henschel2025streamingt2vconsistentdynamicextendable, chen2023videocrafter1opendiffusionmodels, wu2023tuneavideooneshottuningimage, blattmann2023stablevideodiffusionscaling, ma2025lattelatentdiffusiontransformer,  chen2024videocrafter2overcomingdatalimitations, yang2025cogvideoxtexttovideodiffusionmodels, wu2023tuneavideooneshottuningimage, khachatryan2023text2videozerotexttoimagediffusionmodels} and flow models  \cite{hacohen2024ltxvideorealtimevideolatent, Wan2025Wanopenadvancedlargescale, zheng2024opensorademocratizingefficientvideo, kong2025hunyuanvideosystematicframeworklarge, peng2025opensora20trainingcommerciallevel, ma2025stepvideot2vtechnicalreportpractice, polyak2025moviegencastmedia, chen2025skyreelsv2infinitelengthfilmgenerative, jin2025pyramidalflowmatchingefficient} are a class of conditional generative models that synthesize videos from text prompts. Stable Video Diffusion (SVD)  \cite{blattmann2023stablevideodiffusionscaling} focuses on data curation and extends the backbone of Stable Diffusion (SD) \cite{Rombach_2022_CVPR} by temporal layers to capture inter-frame dynamics and improve consistency across the generated frames. SVD faces challenges with scalability, long-term temporal modeling, and T2V alignment. To overcome these, CogVideoX  \cite{yang2025cogvideoxtexttovideodiffusionmodels} employs a 3D variational autoencoder (VAE)  \cite{kingma2022autoencodingvariationalbayes, oord2018neuraldiscreterepresentationlearning, esser2021tamingtransformershighresolutionimage} for spatio-temporal compression and a diffusion transformer (DiT)  \cite{peebles2023scalablediffusionmodelstransformers, bao2023worthwordsvitbackbone} with full 3D attention, where text and visual tokens are jointly processed via self-attention  \cite{vasWani2023attentionneed} to enhance semantic alignment. CogVideoX is computationally expensive due to the compression rate of its VAE. In contrast, LTXV \cite{hacohen2024ltxvideorealtimevideolatent} uses a VAE compression rate four times higher than CogVideoX and pretrains a 2B DiT via flow matching  \cite{esser2024scalingrectifiedflowtransformers, lipman2023flowmatchinggenerativemodeling, albergo2023buildingnormalizingflowsstochastic}. Wan  \cite{Wan2025Wanopenadvancedlargescale} adopts similar architectural and training principles~--~DiT, 3D VAE, flow matching ~--~offering scalable models (1.3B and 14B DiTs) for diverse and high-quality T2V synthesis. 

\textbf{World- and Flow-Based Autoregressive T2V Models for Video Continuation.} Recent methods that generate a video continuation are part of world  \cite{gao2024vista, hassan2024gemgeneralizableegovisionmultimodal} and flow-based autoregressive T2V models  \cite{hacohen2024ltxvideorealtimevideolatent, yin2025causvid}. On the one hand, Vista  \cite{gao2024vista} augments SVD with a latent-replacement scheme that injects up to three historical frames and introduces dynamics-enhancement and structure-preservation losses to improve realism. However, Vista struggles with computational efficiency and controllability. To address the latter, GEM \cite{hassan2024gemgeneralizableegovisionmultimodal} generates the succeeding frames from a reference frame with control over DINOv2  \cite{oquab2023dinov2} features, human poses, and ego-trajectories. GEM adopts progressive noise levels on frames in training to allow generating the next frames. On the other hand, CausVid  \cite{yin2025causvid} distills a bi-directional T2V model into a uni-directional autoregressive student model using asymmetric distribution matching distillation \cite{yin2024onestepdiffusiondistributionmatching, yin2024improveddistributionmatchingdistillation}. It employs a block-wise causal attention along with per-block noise level, preserving bi-directional attention within local blocks while enforcing causality across them. 
Finally, LTXVCondition  \cite{hacohen2024ltxvideorealtimevideolatent} is pre-trained using different amounts of noise levels per-token \cite{chen2024diffusionforcingnexttokenprediction, 
xie2025progressiveautoregressivevideodiffusion, liu2024redefiningtemporalmodelingvideo, kim2024fifodiffusiongeneratinginfinitevideos, ruhe2024rollingdiffusionmodels} allowing generation of a video sequence from a conditioning chunk, which is given to the model as input with zero timestep.



\section{Preliminaries}
\label{sec:sec_2}

\subsection{Flow Matching} 
\label{sec:sec_2_1}

Flow matching  \cite{lipman2023flowmatchinggenerativemodeling, liu2022flowstraightfastlearning, albergo2023buildingnormalizingflowsstochastic} is a training paradigm in generative modeling. It learns to transform a sample from an initial distribution \(p_{init}(x_0)\)~--~typically the standard multivariate normal distribution \(x_0 \sim \mathcal N(0, I_d)\) of data dimensionality \(d\)~--~into \(x_1\) where \(x_1\) comes from the data distribution \(p_{data}\). This is accomplished by first perturbing the target data sample during the training at a given time step \(t\) via a convex combination with noise:

\begin{equation}
    x = tx_1 + (1-t)x_0,
\end{equation}

where \(t \sim  \mathcal U_{[0, 1]}\) is sampled uniformly from the unit interval. The perturbed input \(x\) is then passed to a time-conditioned neural network \(u^\theta_t(x)\). The model is trained to predict the velocity vector \((x_1-x_0)\), using the conditional flow matching loss: 

\begin{equation}
    \mathcal{L_\text{CFM}}(\theta) = ||u^\theta_t(x) - (x_1-x_0)||^2.
\end{equation}

Once \(u^\theta_t(x)\)  has been trained, a new sample from \(p_{data}\) can be generated by first sampling  \(x_0\) from the standard multivariate normal distribution and then following the velocity field predicted by the model using Euler discretization. In the case of linear probability paths, the initial distribution is not restricted to being Gaussian; a flow matching model can be trained between two arbitrary distributions. This motivates us to train a flow matching model between distributions of current and succeeding frames with the goal of generating a video continuation given an input chunk.  

\subsection{Flow Matching with Optimal Couplings} 
\label{sec:sec_2_2}


Multiple strategies exist for choosing a coupling~--~i.e. sampling a pair of \(x_0\) and \(x_1\)~--~during the training (e.g. independent sampling  \cite{lipman2023flowmatchinggenerativemodeling,tong2024improvinggeneralizingflowbasedgenerative} or the use of optimal couplings (OC)  \cite{pooladian2023multisampleflowmatchingstraightening, kornilov2024optimalflowmatchinglearning}). Formally, given the joint distribution \(\Pi ( p_{init}, p_{data})\), the optimal coupling is given by the minimizer of the following optimization problem:

\begin{equation}
\min_{\pi \in \Pi(p_{\text{init}},\, p_{\text{data}})}
\left( \mathbb{E}_{\pi(x_0, x_1)} \left[ \left\| x_0 - x_1 \right\|^2 \right] \right)
\end{equation}

This formulation corresponds to minimizing the squared 2-Wasserstein distance. Employing optimal couplings for sampling pairs of noise and data during training facilitates the learning of straighter flows for \(u^\theta_t(x)\), thereby reducing the number of sampling steps required at inference time to achieve high-quality results in comparison to sampling independently during the training. However, computing optimal couplings in high-dimensional spaces such as images or videos remains computationally intractable in practice. To address this, the authors of \cite{pooladian2023multisampleflowmatchingstraightening, tong2024improvinggeneralizingflowbasedgenerative} propose approximating optimal couplings within a batch at training, demonstrating that such an approach leads to straighter trajectories and a reduced number of inference steps in contrast to non-optimal couplings.

\section{Method}
\label{sec:sec_3}


\subsection{Flowing from Current to Succeeding Frames}
\label{sec:sec_3_1}


Given input frames, the goal of video continuation is to generate the next video chunk that is semantically coherent with the given ones. We propose to flow directly from the input to the succeeding frames using flow matching. Our motivation is twofold. (1) By flowing directly from current to succeeding frames, FlowC2S avoids the conventional approach~--~such as LTXVCondition  \cite{hacohen2024ltxvideorealtimevideolatent} and CausVid  \cite{yin2025causvid}~--~of providing  both given frames and additional noise at inference, thereby halving the input dimensionality. (2) We hypothesize that learning a vector field between current and succeeding frames may facilitate faster convergence and enable more efficient inference-time sampling compared with mapping from noise to future frames. Formally, given \(x_0\) from the distribution of the current frame chunks \(p_\text{c\_fc}\) and \(x_1\) from the distribution of succeeding frame chunks \(p_\text{s\_fc}\) our goal is to learn a parametric vector field \(u^\theta_t(x)\) to predict the velocity between the input and succeeding chunks. In practice, flow matching is performed in a pre-trained 3D VAE's  \cite{kingma2022autoencodingvariationalbayes} latent space. Algorithm~\ref{algo:algo-1} presents our training algorithm.

\begin{algorithm}[t]
\caption{Flowing From Current To Succeeding Frames}
\begin{algorithmic}[1]
\State \textbf{Require:} \text{pretrained} $u_{t}^{\theta^{\text{*}}}$, \(\Pi ( p_{\text{c\_fc}}, p_{\text{s\_fc}})\), $\rho$
\State $u_t^{\theta} \gets u_t^{\theta^*}$
\For{$x_0, x_1 \sim \Pi$ \textbf{with} $(x_0, x_1)$ \textbf{inherent optimal couplings}}
    \State $\mu_1, \sigma_1 = \text{VAE}(x_1), \quad x_1 \gets \mu_1$
    \State $\mu_0, \sigma_0 = \text{VAE}(x_0)$
    \LineComment{Compute inverted latent}
    \State $\hat{x}_1 = $ \text{RF-Solver-Inversion}($u_t^{\theta^*}, x_1$)  \cite{Wang2025tamingrectifiedflowinversion}
    \LineComment{Apply Target Inversion}
    \State \textbf{if} $p < \rho$ \textbf{then} $x_0 \gets \mu_0 + \sigma_0 \hat{x}_1$ \textbf{else} $x_0 \gets \mu_0$ 
    \State $t \sim \mathcal{U}[0, 1]$
    \State $x \gets (1 - t)x_0 + tx_1$
    \State $\mathcal{L}_{\text{CFM}}(\theta) = \left\| u_t^\theta(x) - (x_1 - x_0) \right\|^2$
    \State Update $\theta$ using GD on $\mathcal{L}_{\text{CFM}}(\theta)$
\EndFor
\end{algorithmic}
\label{algo:algo-1}
\end{algorithm}

\subsection{Inherent Optimal Couplings}
\label{sec:sec_3_2}

As discussed in Section~\ref{sec:sec_2_2}, learning a vector field with optimal couplings leads to improved convergence and straighter sample trajectories. However, in high-dimensional settings such as videos, computing exact optimal couplings under a cost function is computationally intractable. Thus, we propose a practical and structure-aware approximation: treating temporally adjacent video chunks of the same video as inherently optimal or near-optimal couplings. Specifically, we partition a given video into consecutive segments: the first segment serves as the input chunk \(x_0\), and the second as the succeeding chunk \(x_1\). This pairing leverages the natural temporal continuity and semantic coherence within a single video, which makes it significantly more consistent than random pairings of chunks across different videos. Crucially, such intra-video couplings incur no additional computational overhead, yet act as a strong proxy for true optimal transport couplings. Empirically, we observe that using these temporally ordered pairs at training results in decreased loss values (Fig.~\ref{fig:fig_5} (left)) and reduces the number of neural function evaluations required for high-quality inference in comparison to independent pairs (Fig.~\ref{fig:fig_4_a}). This observation is consistent with prior findings that straighter flows can be achieved when optimal or approximate-optimal couplings are used  \cite{pooladian2023multisampleflowmatchingstraightening, tong2024improvinggeneralizingflowbasedgenerative}.

\begin{figure*}[t]
    \centering
    \subfloat[
    Disallowing a video chunk from matching with itself.
    ]{%
        \includegraphics[width=0.45\linewidth]{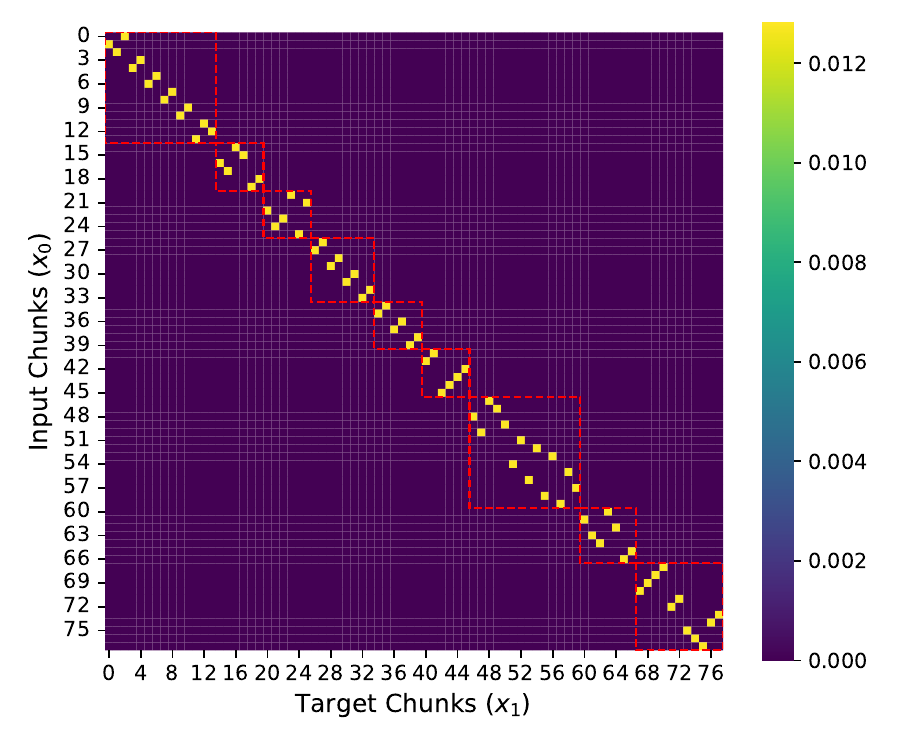}%
        \label{fig:fig_1_a}
    }\hfill
    \subfloat[Matching only with succeeding chunks across videos.]{%
        \includegraphics[width=0.45\linewidth]{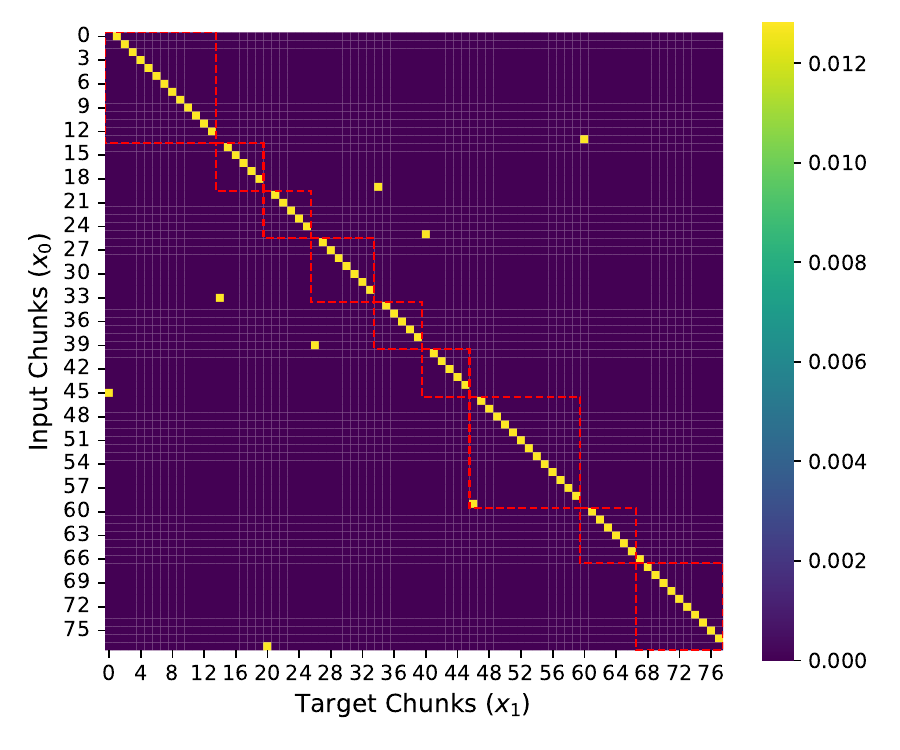}%
        \label{fig:fig_1_b}
    }
    \caption{Optimal Transport (OT) plan heatmaps between video chunks. We compute pairwise OT plans between a batch of video chunks, solving the Monge–Kantorovich OT problem with uniform marginals. Red dashed lines mark individual video boundaries. The structure of the transport plans reveals strong alignment between temporally adjacent chunks from the same video, even without explicit temporal constraints in (b).}
    \label{fig:fig_1}
\end{figure*}

To empirically validate our hypothesis that consecutive chunks from the same video approximate optimal couplings, we compute pairwise Optimal Transport (OT) plans between a batch of video chunks. Specifically, we randomly sample ten videos from our training set, OpenVid  \cite{nan2024openvid}, and divide each into chunks of 41 frames. We then solve the Monge–Kantorovich OT problem  \cite{Kantorovich1948_Monge} subject to uniform marginals:


\begin{equation}
\min_{\pi \in \mathbb{R}^{n \times n}} \;
\sum_{i=1}^{n} \sum_{j=1}^{n} M_{ij} \, \pi_{ij},
\label{eq:eq_4}
\end{equation}

such that \(\sum_{j=1}^{n} \pi_{ij} = \tfrac{1}{n}, \;\; \forall i,
\sum_{i=1}^{n} \pi_{ij} = \tfrac{1}{n}, \;\; \forall j \), where \(\pi_{ij}\) indicates the probability of matching source  \(i\) with target  \(j\), \(n\) refers to the number of video chunks and \( M_{ij} \) denotes the pairwise squared Euclidean cost between the \( i \)-th and \( j \)-th video chunks. Fig.~\ref{fig:fig_1} presents heatmaps of the resulting OT plans under two different masking strategies: in Fig.~\ref{fig:fig_1_a}, self-matching is disallowed (i.e., a video chunk can not match with itself), whereas in Fig.~\ref{fig:fig_1_b}, each chunk is permitted to match only with succeeding chunks across videos. Red dashed lines in Fig.~\ref{fig:fig_1} demarcate the boundaries between individual videos. The spurious matches observed for the final chunk of each video in Fig.~\ref{fig:fig_1_b} are natural artifacts arising from the absence of valid succeeding chunks. The overall coupling patterns exhibit a strong preference for temporally adjacent segments within the same video; even in the absence of explicit temporal constraints, suggesting that such chunk pairs can serve as structure-aware, computation-free approximations of optimal couplings.

\subsection{Target Inversion} 
\label{sec:sec_3_3}

Replacing the multivariate normal distribution with the empirical distribution of the current frames when fine-tuning a pre-trained T2V model also changes the generative objective faced by the pre-trained vector field, since the target vector field becomes the difference between succeeding and current chunks. To bridge this mismatch, we introduce Target Inversion (TI): given a chunk of succeeding frames  \(x_1\), we recover an inverted latent \(\hat{x}_1\) using RF-Solver inversion  \cite{Wang2025tamingrectifiedflowinversion} and the pre-trained model  such that the Euler solver reconstructs  \(x_1\) from \(\hat{x}_1\). Then, during training, with probability \(\rho\), we leverage \(\hat{x}_1\) when sampling a latent representation for the input frames. We hypothesize that this biases the model toward learning the residual shift-and-scale transformation  applied to the inverted target latent. TI yields notable gains in user preference (\cref{tab:user_study}). TI analysis by motion speed and camera movement is in the appendix.

\begin{table*}[t]
\centering
\begin{tabular}{l ccc ccc ccc}
\toprule
\multirow{3}{*}{Method}
  & \multicolumn{6}{c}{Visual Quality} & \multicolumn{3}{c}{Efficiency} \\
\cmidrule(lr){2-7}\cmidrule(lr){8-10}
  & \multicolumn{3}{c}{OpenVid} & \multicolumn{3}{c}{NuScenes}
  & \multirow{2}{*}{\shortstack{Total \\ NFE $\downarrow$}}
  & \multirow{2}{*}{$k$ (MB\,/\,$10^6$) $\downarrow$}
  & \multirow{2}{*}{$b$ (MB)} \\
\cmidrule(lr){2-4}\cmidrule(lr){5-7}
  & FID $\downarrow$ & FVD $\downarrow$ & $\mathrm{LPIPS_s}$ $\downarrow$
  & FID $\downarrow$ & FVD $\downarrow$ & $\mathrm{LPIPS_s}$ $\downarrow$
  & & & \\
\midrule
\multicolumn{10}{c}{\textit{Backbone: SVD\,(1.5B)}} \\
\midrule
Vista \cite{gao2024vista}
  & 0.38 & 184.33 & 0.37
  & 3.50 & 368.08 & 0.49
  & 50  & 20244.37 & 18\,430.21 \\
GEM \cite{hassan2024gemgeneralizableegovisionmultimodal}
  & 1.47 & 413.60 & 0.55
  & 1.89 & 251.13 & 0.61
  & 117
  & 6934.35 & 4194.67 \\
\midrule
\multicolumn{10}{c}{\textit{Backbone: LTXV\,(2B)}} \\
\midrule
LTXVCond. \cite{hacohen2024ltxvideorealtimevideolatent}
  & \underline{0.46} & 134.13 & 0.30
  & 1.75 & 228.72 & 0.44
  & 40
  & \underline{7103.22} & $-$660.92 \\
\multirow{2}{*}{Ours}
  & 0.48 & \underline{129.20} & \underline{0.23}
  & \underline{1.13} & \underline{185.48} & \underline{0.41}
  & \textbf{5}
  & \multirow{2}{*}{\textbf{3552.32}} & \multirow{2}{*}{$-$308.35} \\
  & \textbf{0.44} & \textbf{124.97} & \textbf{0.22}
  & \textbf{1.08} & \textbf{172.86} & \textbf{0.40}
  & \underline{10}
  & & \\
\midrule
\multicolumn{10}{c}{\textit{Backbone: Wan\,(1.3B)}} \\
\midrule
CausVid \cite{yin2025causvid}
  & 0.30 & 379.16 & 0.32
  & 1.28 & 812.46 & 0.43
  & \textbf{5}
  & \underline{93.35} & 4003.58 \\
\multirow{2}{*}{Ours}
  & \underline{0.10} & \underline{105.51} & \underline{0.14}
  & \underline{0.45} & \underline{158.97} & \underline{0.32}
  & \textbf{5}
  & \multirow{2}{*}{\textbf{48.64}} & \multirow{2}{*}{3998.74} \\
  & \textbf{0.06} & \textbf{98.46} & \textbf{0.13}
  & \textbf{0.33} & \textbf{125.63} & \textbf{0.31}
  & \underline{10}
  & & \\
\bottomrule
\end{tabular}
\caption{Quantitative comparison on OpenVid and NuScenes (val.). Lower is better ($\downarrow$). Best and second-best scores are bolded and underlined, respectively, within each backbone group. Total NFE: the number of forward evaluations. LTXVCond.: LTXVCondition. GPU memory scaling is approximated by $\mathrm{Mem}(V)\!\approx\!k\cdot(V/10^6)+b$ (OLS); smaller $k$ is better.}
\label{tab:tab_2}
\end{table*}
\section{Experiments}
\label{sec:sec_4}

\subsection{Experimental Setup}
\label{sec:sec_4_0}

We evaluate FlowC2S on two backbones~\cite{hacohen2024ltxvideorealtimevideolatent, Wan2025Wanopenadvancedlargescale} and two datasets~\cite{nan2024openvid, caesar2020nuscenesmultimodaldatasetautonomous}, comparing against world (Vista~\cite{gao2024vista}, GEM~\cite{hassan2024gemgeneralizableegovisionmultimodal}) and autoregressive T2V methods (LTXVCondition~\cite{hacohen2024ltxvideorealtimevideolatent}, CausVid~\cite{yin2025causvid}). 


\textbf{Datasets.} We train our models on a randomly selected subset of 400K videos from OpenVid \cite{nan2024openvid}, a dataset of high-quality and diverse video clips. To mitigate abrupt scene transitions in videos, we apply a histogram-based scene change detector to segment videos into coherent clips. Additionally, in preparing the training dataset, we sample only non-overlapping, consecutive video segments from each video to prevent any chunk from appearing in both the source and target distributions at training. For evaluation, we use a randomly sampled set of 2000 videos from OpenVid and another random sample of 2000 videos from the NuScenes  \cite{caesar2020nuscenesmultimodaldatasetautonomous} validation set used by Vista  \cite{gao2024vista}.


\textbf{Evaluation Metrics.} We analyze the number of total neural function evaluations (NFEs) and GPU memory overhead as measures of efficiency. The total NFE reflects the number of network passes through the diffusion (or flow) backbone. The quality of the generated videos is evaluated using the Fréchet Inception Distance (FID) \cite{heusel2018ganstrainedtimescaleupdate}, the Fréchet Video Distance (FVD) \cite{unterthiner2019accurategenerativemodelsvideo} against the ground truth chunk, as well as LPIPS$_\text{s}$ \cite{zhang2018perceptual} at the input-output transition boundary to measure seam smoothness. For a fair comparison, the generated and ground-truth chunks are ensured to contain the same number of frames in all experiments. For evaluation, we generate video continuations given an input video chunk. Owing to input-frame constraints in Vista and the NuScenes validation set, we use 17 input frames and generate the subsequent 17 frames. For ablation studies on the method and NFE, we set the number of current and succeeding frames to 41 and use a spatial resolution of \(256\times 384\). In addition, we report a subset of VBench metrics \cite{huang2023vbench} in the appendix, where we also provide more details on the evaluation protocol.


\textbf{Implementation Details.} We fine-tune our models initialized from LTXV (2B) and Wan (1.3B) on H100 GPUs with a total batch size of 14,336 for 1406 steps, using AdamW  \cite{loshchilov2019decoupledweightdecayregularization} with a learning rate of 
\(2\times10^{-4}\). We use 17 and 41 frame video chunks at training, spatially resized to 
\(256\times384\) when using LTXV as a backbone, and to \(240\times416\) when using Wan as a backbone. All training hyperparameters are in the appendix.

\begin{figure*}[t]
    \centering
    \includegraphics[width=1.0\linewidth]{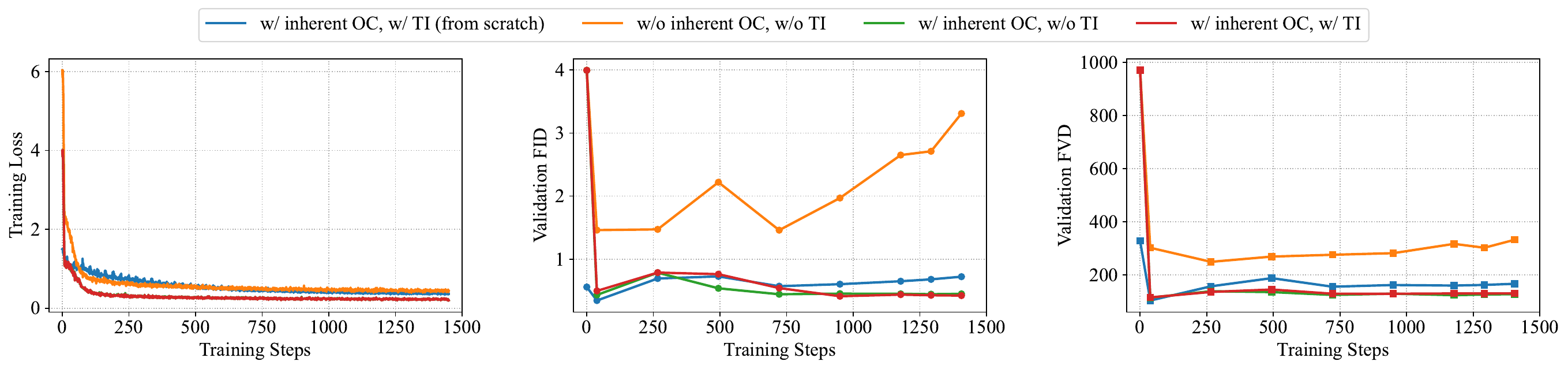}%
    \caption{Training loss (left), validation FID (middle), and FVD (right) across four experimental set-ups. Training from scratch with inherent Optimal Couplings (OC) and Target Inversion (TI) is ineffective in terms of training loss. Fine-tuning without inherent OC+TI yields worse FID and FVD than all other settings. Fine-tuning with inherent OC+TI attains the lowest loss and best validation FID and FVD across configurations; relative to inherent OC-only fine-tuning, quantitative scores are comparable, but inherent OC+TI yields better visual quality.}
    \label{fig:fig_5}
\end{figure*}

\begin{figure*}[t]
    \centering
    \includegraphics[width=1.0\linewidth]{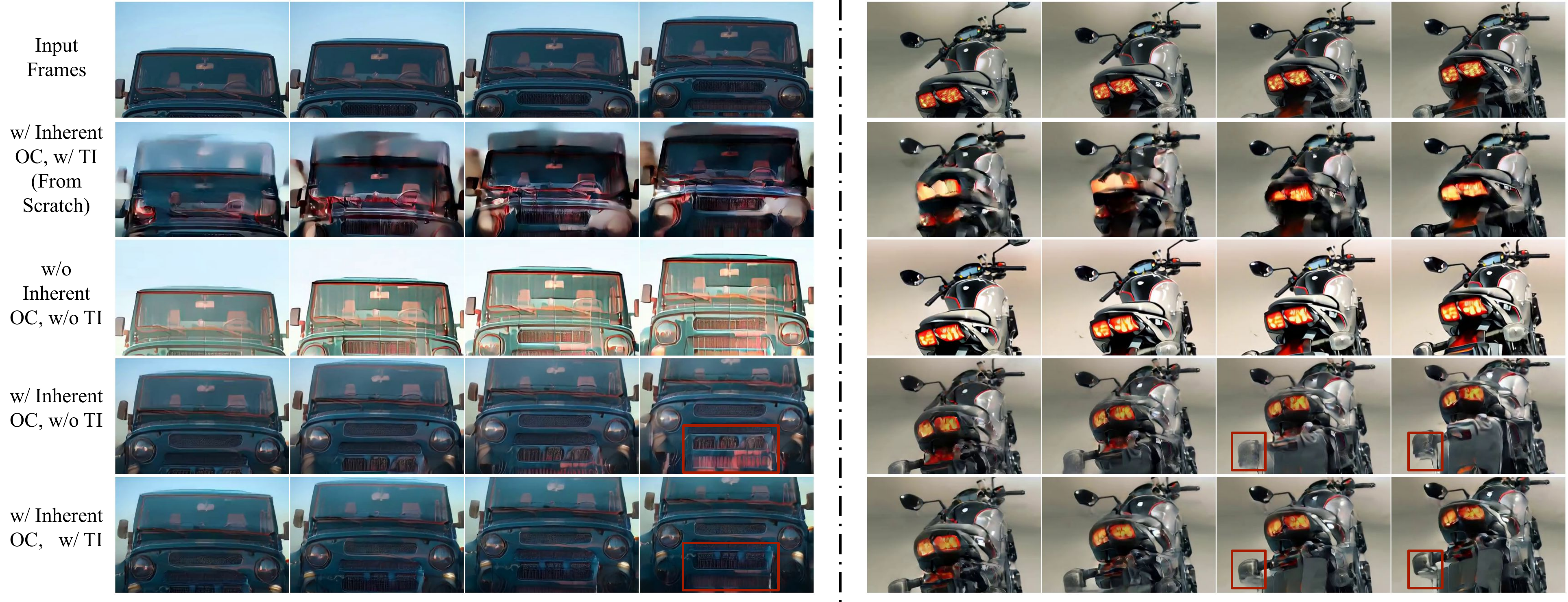}%
    \caption{Visual comparison across four settings; frames shown with a stride of 13. Training from scratch w/ OC+TI shows visual artifacts, indicating poor convergence. Fine-tuning w/o OC+TI yields temporal incoherence and color shifts. Fine-tuning w/ OC+TI produces sharper, more consistent details than OC-only fine-tuning (see red rectangles). Overall, the w/ OC+TI setting achieves the best visual quality; more in appendix.}
    \label{fig:fig_6}
\end{figure*}

\subsection{Comparison with State-of-the-Art}
\label{sec:sec_4_4}



\begin{figure*}[t]
    \centering
    \subfloat[Ablation on the number of function evaluations (NFE).]{%
        \includegraphics[width=0.3\linewidth]{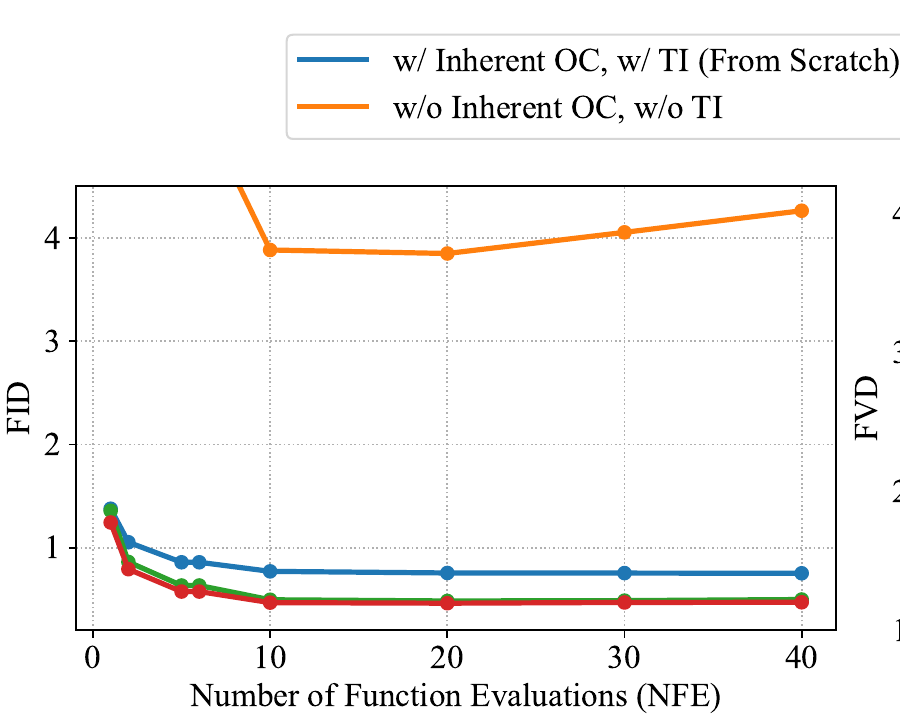}%
        \includegraphics[width=0.275\linewidth]{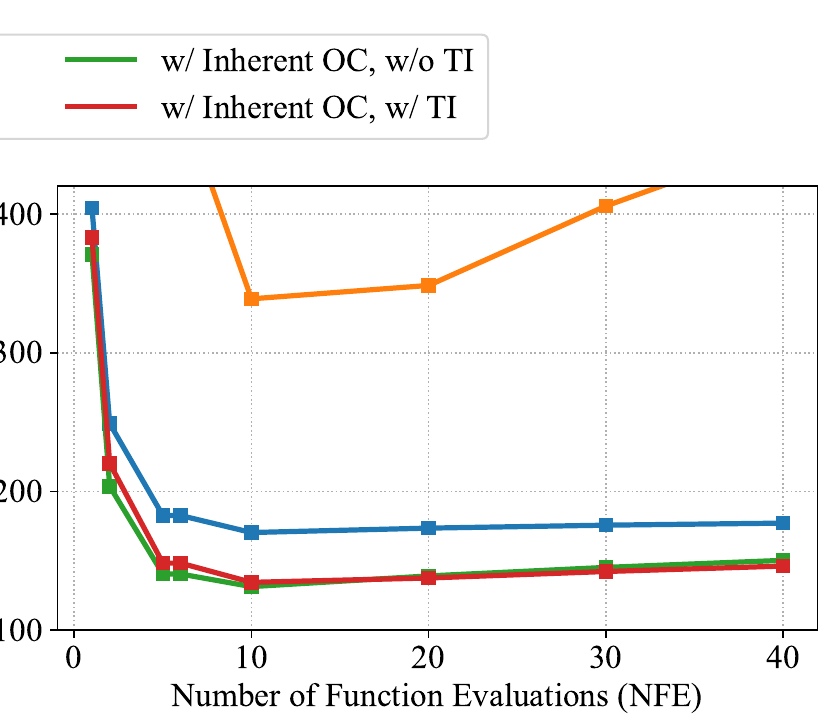}%
        \label{fig:fig_4_a}
    }\hfill
    \subfloat[Ablation on number of frames.]{%
        \includegraphics[width=0.325\linewidth]{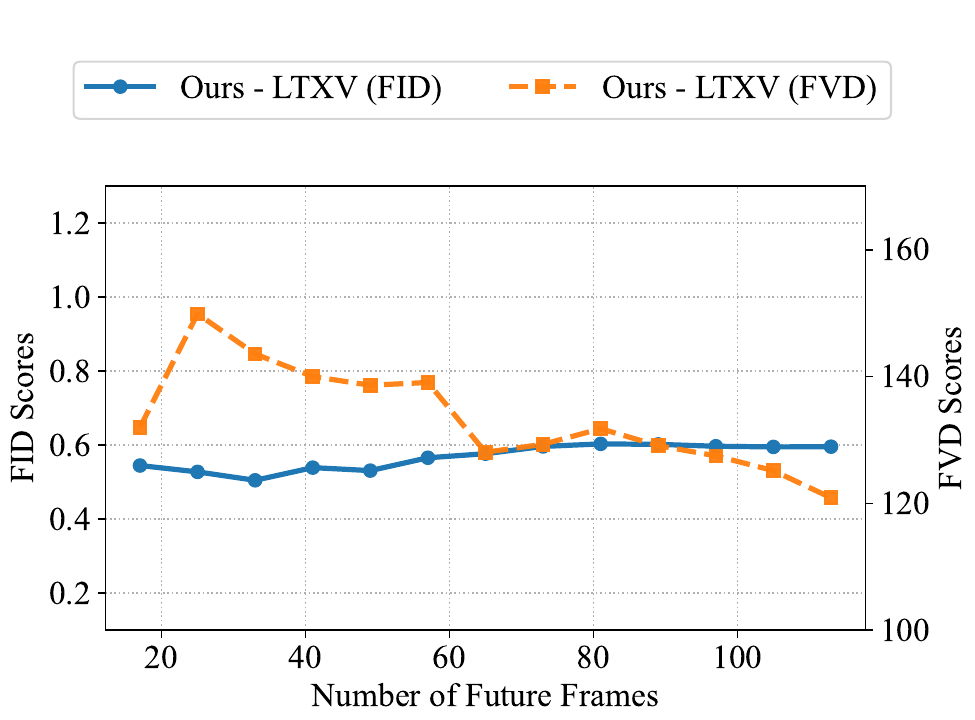}%
        \label{fig:fig_4_b}
    }
    \caption{Ablations on NFE and number of frames: (a) With inherent OC+TI, 5–10 NFEs equate or surpass 40 NFEs on FID/FVD and outperform fine-tuning without inherent OC/TI, indicating FlowC2S preserves high visual and temporal quality with fewer NFEs; (b) FlowC2S keeps FID/FVD stable as number of generated frames in video continuation grows, generalizing beyond its 17/41-frame training.}
    \label{fig:fig_4}
\end{figure*}

\textbf{Efficiency and Quantitative Results.} Table~\ref{tab:tab_2} provides a quantitative comparison between the state-of-the-art and FlowC2S (fine-tuned from LTXV and Wan) on the OpenVid and NuScenes validation sets and the total number of neural function evaluations (NFE) per method. For CausVid, NFE depends on the size of the KV cache, which we set equal to the number of input frames (i.e. 17). Our LTXV-based model achieves superior FID and FVD scores compared to LTXVCondition, while requiring four times fewer NFEs. Likewise, our Wan-based model outperforms CausVid at similar NFEs and has half the input model dimensionality. Moreover, our Wan-based model attains these quantitative results using a straightforward flow matching loss, without additional distillation as required by CausVid.  

Table~\ref{tab:tab_2} also presents an analysis of GPU memory overhead during generation in comparison to the state-of-the-art. We define the effective volume \(V\) as the product of the latent input tensor dimensions, \(c\times f\times h\times w \) where \(c\) is the number of channels, \(f\) the number of frames, and \(h\) and \(w\) the spatial resolution. For each method, we evaluate the average end-to-end GPU overhead (including VAE encoding and decoding) over 100 runs at multiple effective volumes. As different methods employ different VAE compression ratios, we ensure that the effective volumes are matched across models for a fair comparison, see the appendix for details. To characterize scaling behavior, we assume a linear relationship of the form \(\text{Mem} (V)\!\approx\!k\cdot(V/10^6)+b\) where \(k\) denotes the slope (memory increase per million effective volume units) and \(b\) the intercept (constant overhead). \(k\) and \(b\) are estimated using ordinary least squares (OLS) regression. We compare our LTXV-based model to LTXVCondition and our Wan-based model to CausVid. In both cases, FlowC2S achieves almost a twofold reduction in  \(k\), aligning with our goal of reducing input dimensionality.

\textbf{Qualitative Results.} Fig.~\ref{fig:fig_teaser} presents visualizations of video continuations on OpenVid (val.), generated by our LTXV-based model with total NFE of \(10\) at a resolution of \(256\times 384\) with \(41\) current and succeeding frames. The generated videos exhibit detailed motion, maintain temporal coherence without explicit conditioning on the input frames, and yield logically plausible continuations of the given input video. Observe the gradual camera zoom-out in the sand-and-rock scene, which our method successfully generates. Similarly, notice how the boat moves from right to left, a motion that continues in the generated frames by our method. Additional visual results are in the appendix.

\subsection{Ablation Studies}
\label{sec:sec_4_5}

\textbf{Effect of Fine-tuning, Inherent OC and TI.} Fig.~\ref{fig:fig_5} ablates the principal design choices of FlowC2S: fine-tuning from a pre-trained T2V flow model (LTXV) rather than training from scratch;  using inherent optimal couplings (OC) as a proxy for the true optimal couplings at training; and employing target inversion (TI). We compare training losses and validation metrics (FID, FVD) across four configurations: (1) training from scratch with inherent OC+TI, (2) fine-tuning without inherent OC and without TI, (3) fine-tuning with inherent OC but without TI, and (4) fine-tuning with inherent OC+TI (Algorithm~\ref{algo:algo-1}). In training from scratch, we initialize the DiT of LTXV with random weights. The implementation details of (2) and (3) are provided in the appendix. Three main observations emerge. First, training from scratch with inherent OC+TI yields higher losses than fine-tuning from LTXV with inherent OC+TI, underscoring the importance of leveraging a pre-trained model's prior knowledge about video generation for video continuation. Although at step 38 the model trained from scratch with inherent OC+TI reports lower FID and FVD scores than the other configurations, this effect is misleading. At such an early stage of training, the model has not yet learned meaningful dynamics and instead tends to generate results with minimal variation from the given input, which artificially improves FID and FVD when compared against the ground-truth segment. However, visually, the outputs in \cref{fig:fig_6} reveal clear degradation, including interpolation artifacts. Second, omitting inherent OC and TI at fine-tuning increases losses and degrades FID and FVD relative to fine-tuning with inherent OC+TI. Although fine-tuning without OC+TI shows losses similar to training from scratch with inherent OC+TI, qualitative inspection reveals generated frames that are inconsistent with the conditioning frames (Fig.~\ref{fig:fig_6}); moreover, continuing fine-tuning with inherent OC+TI beyond step 38 further improves validation FID. Finally, fine-tuning with inherent OC+TI and with inherent OC-only attains comparable losses and FID and FVD, yet inherent OC+TI yields visibly higher perceptual quality in the generated videos (Fig.~\ref{fig:fig_6}).

\textbf{Ablation on NFE. } Fig.~\ref{fig:fig_4_a} reports a quantitative ablation over NFEs for FlowC2S fine-tuned from LTXV across the four experimental settings indicated above (see the appendix for visual results). Using 5–10 NFEs matches or outperforms the 40-NFE setting on both FID and FVD when fine-tuning with inherent OC and TI, demonstrating that high perceptual quality and temporal coherence can be achieved with notably fewer sampling steps. Excluding inherent OC and TI results in substantially worse FID and FVD at \(\text{NFE}=40\) than fine-tuning with inherent OC and TI at \(\text{NFE}=10\). This supports our design choice that leveraging inherent optimal couplings at training enables efficient inference without sacrificing fidelity. 

\textbf{Ablation on Long Video Continuation.} Fig.~\ref{fig:fig_4_b} presents a quantitative analysis of performance as a function of \# generated frames with visuals in \cref{fig:fig_teaser} and the appendix. FlowC2S, fine-tuned from LTXV, maintains consistent FID and FVD as \# generated frames grows~--~despite being trained only on 17- and 41-frame video chunks~--~indicating stable per-frame fidelity, temporal coherence, and effective generalization to the generation of long video continuations across the evaluated ranges.

\begin{table}[t]
\centering
\begin{tabular}{l r r r}
\toprule
Metric & w/ TI & Similar & w/o TI \\
\midrule
Overall Preference  & \textbf{47.5} & 36.2 & 16.2 \\
Temporal Coherence  & \textbf{47.5} & 34.4 & 18.1 \\
Visual Quality      & \textbf{49.4} & 35.0 & 15.6 \\
\bottomrule
\end{tabular}
\caption{Human preference study comparing results with and without Target Inversion (TI). Percentages indicate the fraction of pairwise comparisons in which each method was preferred.}
\label{tab:user_study}
\end{table}

\textbf{User Study on TI.} To validate the perceptual quality improvement with TI, we conducted a user study with 15 participants comparing outputs with inherent OC, without TI and with inherent OC, with TI across 20 videos. Participants were shown the input video along with two results—one with and one without TI—in randomized order. The video order was also randomized among participants. Users evaluated visual quality, temporal coherence, and overall preference. For each question, participants could choose either of the generated videos or indicate a similar quality. As shown in \cref{tab:user_study}, users exhibited a three times stronger preference for outputs with TI.


\newlength{\plotsgap}
\setlength{\plotsgap}{0.02\linewidth} 

\newlength{\axw}
\setlength{\axw}{\dimexpr(\linewidth - \plotsgap)/3\relax}

\section{Limitations}

Our training dataset is curated to exclude scene transitions, limiting the model’s ability to generate complex motions and abrupt scene changes. In addition, all training experiments use fixed-length chunks (17 and 41 frames). The method empirically generalizes to longer videos but does not yet model variable-length inputs. Future work includes training with heterogeneous chunk lengths and adding controllability via explicit conditioning (e.g., rendering maps, camera trajectories) \cite{
zhang2023addingconditionalcontroltexttoimage, 
zhang2023controlvideotrainingfreecontrollabletexttovideo,
jiang2025vaceallinonevideocreation, he2025cameractrlenablingcameracontrol, 
xu2024camcocameracontrollable3dconsistentimagetovideo, zheng2025vidcraft3cameraobjectlighting,  ma2025controllablevideogenerationsurvey} for user-directed dynamics. Further limitations and directions are in the appendix.
\section{Conclusion}

This paper introduced FlowC2S, a video continuation approach that directly flows from input to succeeding frames. Our method fine-tunes text-to-video priors (e.g., LTXV/Wan) from the distribution of current to succeeding frames, adopting inherent optimal couplings via adjacent chunks, and integrating target inversion into the training. Empirically, FlowC2S achieves state-of-the-art FID and FVD scores with as few as five neural function evaluations, while delivering a \(2\times\) reduction in model input dimensionality relative to the state-of-the-art.

\section*{Acknowledgments}

This work was granted access to the HPC resources of IDRIS under the allocations 2023-AD011014851, 2025-A0191016617, 2025-AD011016617R1 and 2025-AD011016643 made by GENCI, and supported by French government funding managed by the National Research Agency (ANR) under the Investments for the Future program (PIA) with the grant ANR-22-EXEN-0002 (eNSEMBLE project). The authors would like to thank Arijit Ghosh and Vicky Kalogeiton for their helpful insights and fruitful discussions.


{
    \small
    \bibliographystyle{ieeenat_fullname}
    \bibliography{main}
}

\clearpage
\maketitlesupplementary

\renewcommand{\thesection}{A\arabic{section}}
\setcounter{section}{0}

\begin{table*}[t]
\centering
\begin{tabular}{lcc}
\toprule
\textbf{Configuration} & \textbf{LTXV-based} & \textbf{Wan-based} \\
\midrule
Batch Size / GPU & 64 & 32 \\
Accumulate Step & 8 & 8 \\
Optimizer & AdamW & AdamW \\
$\beta_1$ & 0.9 & 0.9 \\
$\beta_2$ & 0.99 & 0.99 \\
Learning Rate & 0.0002 & 0.0002 \\
Learning Rate Schedule & Linear & Cosine \\
Training Steps & 1450 & 1450 \\
Resolution & 256$\times$384 & 240$\times$416 \\
Number of Frames & 17, 41 & 17, 41 \\
Shifting & True & True \\
Weighting Scheme & Logit Normal & Uniform \\
Num Layers & 28 & 30 \\
$p$ & 0.7 & 0.7 \\
Pre-trained Model & LTX-Video-2b-v0.9.5 & Wan2.1-T2V-1.3B \\
\midrule
Sampler & \shortstack{FlowMatchEulerDiscreteScheduler\\ \cite{esser2024scalingrectifiedflowtransformers}} & \shortstack{UniPCMultistepScheduler\\ \cite{zhao2023unipcunifiedpredictorcorrectorframework}} \\
Sample Steps & 40 & 50 \\
Classifier-Free Guidance Scale & 3.5 & 5 \\
\midrule
Device & NVIDIA H100 80 GB $\times$28 & NVIDIA H100 80 GB $\times$56 \\
Training Strategy & AMP / DDP / BFloat16 & AMP / DDP / BFloat16 \\
\bottomrule
\end{tabular}
\caption{Fine-tuning hyper-parameters used in our experiments.}
\label{tab:hyperparams}
\end{table*}


This supplementary material presents additional implementation details of FlowC2S with further details on the ablation studies, offers extended visual results, and concludes with a discussion of its limitations.

In Appendix~\ref{appendix:imp_details}, we describe the implementation of the method, evaluation setup, the protocol used for the NuScenes dataset \cite{caesar2020nuscenesmultimodaldatasetautonomous}, and details of the analysis on the effective GPU memory volume. 

Appendix~\ref{appendix:eval_vbench} reports additional quantitative results; in particular, evaluation  using VBench \cite{huang2023vbench} metrics on OpenVid \cite{nan2024openvid} and NuScenes \cite{caesar2020nuscenesmultimodaldatasetautonomous}, comparison with Bi-Flow \cite{Liu_2025_ICCV}, and the effect of video chunk size during training.

Appendix~\ref{appendix:appendix_more_algos} provides the algorithms used in the ablation studies, complementing the main algorithm described in the paper. 

In Appendix~\ref{appendix:more_vis_results}, we report additional visual results for both our method and the ablation variants. 

Finally, Appendix~\ref{appendix:lim_and_future_work} discusses the limitations of the proposed method and outlines potential directions for future work.

\section{Implementation and Evaluation Details}
\label{appendix:imp_details}


\subsection{Training Details}

The proposed FlowC2S method is fine-tuned from two text-to-video backbones: LTXV \cite{hacohen2024ltxvideorealtimevideolatent} and Wan \cite{Wan2025Wanopenadvancedlargescale}. All ablation studies are run exclusively on LTXV as LTXV’s VAE \cite{kingma2022autoencodingvariationalbayes} provides a higher overall compression ratio than Wan, yielding faster training and lower compute/memory cost under the same hardware budget (i.e., for the Wan backbone, we double the number of devices relative to LTXV to ensure a matched batch size.).  

Unless stated otherwise, we use the AdamW optimizer \cite{loshchilov2019decoupledweightdecayregularization} with a learning rate of 0.0002. The complete set of hyperparameters used for fine-tuning from both LTXV and Wan is listed in Table~\ref{tab:hyperparams}.

\subsection{Evaluation Setup}

We evaluate the proposed FlowC2S approach for generating video continuations against world \cite{gao2024vista, hassan2024gemgeneralizableegovisionmultimodal} and autoregressive \cite{hacohen2024ltxvideorealtimevideolatent, yin2025causvid} text-to-video methods in a video continuation setting with a fixed number of frames: each model receives 17 conditioning (or input) frames and generates the next 17 frames. The 17/17 choice is driven by the maximum sequence length that fits in memory for Vista \cite{gao2024vista} (i.e., 34 total frames per sample) on a single NVIDIA H100 GPU with 80 GB of memory. To ensure a fair comparison, we adopt the same 17-frame input and 17-frame output for all methods, including ours. Table~\ref{tab:method_configs} reports all evaluation hyperparameters across methods (e.g., input/generation resolution, classifier-free guidance scale \cite{ho2022classifierfreediffusionguidance}, and other runtime settings).

\subsubsection{NuScenes Protocol}

For the NuScenes dataset \cite{caesar2020nuscenesmultimodaldatasetautonomous}, we use the validation split from Vista (150 scenes; 750 videos per camera position). We aggregate three camera views~--~FRONT, BACK, and FRONT-LEFT~--~and randomly sample 2,000 videos for evaluation to match the sample count of our OpenVid \cite{nan2024openvid} validation set. The exact validation indices for OpenVid and NuScenes are provided with this supplementary material.

\begin{table*}[t]
\begin{tabular}{lccccccccc}
\\ \toprule
Method & H & W & In &
\shortstack{Out\\(img/lat)} &
Out-L &
\shortstack{Blk\\(frames)} &
\shortstack{Total\\NFE} &
CFG &
Rnd \\
\midrule
Vista          & 576 & 1024 & 17 & 17 & 17 & N.A. & 50      & 2.5 & 1 \\
GEM            & 576 & 576  & 17 & 17 & 17 & N.A. & 117     & 1.5 & 1 \\
CausVid        & 480 & 832  & 17 & 17 & 5  & 5 & 5       & 1   & N.A. \\
LTXVCondition  & 256 & 256  & 17 & 17 & 3  & N.A. & 40      & 3.5 & N.A. \\
Ours (LTXV)    & 256 & 384  & 17 & 17 & 3  & N.A. & 5/6/10  & 3.5 & N.A. \\
Ours (Wan)     & 240 & 416  & 17 & 17 & 5  & N.A. & 5/6/10  & 5   & N.A. \\
\bottomrule
\end{tabular}
\centering
\caption{Evaluation hyper-parameters across all methods being compared. H: Height, W: Width, In: the number of conditioning or input frames, Out: the number of output frames, Out-L: the number of output frames in the latent space, Blk: frames per block, NFE: number of function evaluations, CFG: classifier-free guidance scale, Rnd: number of sampling rounds, N.A.: not applicable.}
\label{tab:method_configs}
\end{table*}
\vspace*{2em}

\subsection{Details on Effective Volume for GPU Memory Analysis}
\label{appendix:details_mem_eff}

We compare GPU memory usage across world and autoregressive text-to-video models that rely on different backbones and therefore different VAE compression ratios. To ensure a fair comparison across all methods, we match (or closely approximate) the effective volume per run and per method. Vista and GEM \cite{hassan2024gemgeneralizableegovisionmultimodal} use the SVD \cite{blattmann2023stablevideodiffusionscaling} backbone; LTXVCondition uses LTXV \cite{hacohen2024ltxvideorealtimevideolatent}; CausVid \cite{yin2025causvid} uses Wan \cite{Wan2025Wanopenadvancedlargescale}; our method is fine-tuned and evaluated with LTXV and Wan backbones. 

\begin{table}[t]
\centering
\begin{tabular}{r rr ccc}
\\ \toprule
\multicolumn{1}{c}{\#} &
\multicolumn{1}{c}{$V_{\text{SVD}}$} &
\multicolumn{1}{c}{$V_{\text{LTXV/Wan}}$} &
\multicolumn{1}{c}{$F_\text{SVD}$} &
\multicolumn{1}{c}{$F_\text{LTXV}$} &
\multicolumn{1}{c}{$F_\text{Wan}$ } \\
\midrule
1 & 294{,}912 & 299{,}520 & 5   & 41   & 9   \\
2 & 368{,}640 & 399{,}360 & 8  & 57  & 13  \\
3 & 479{,}232 & 499{,}200 & 10 & 73  & 17  \\
4 & 589{,}824 & 599{,}040 & 13 & 89  & 21 \\
5 & 884{,}736 & 898{,}560 & 16 & 137  & 33 \\
\bottomrule
\end{tabular}
\caption{Representative data used in the GPU–memory analysis across five effective volumes. \(V_\text{SVD}\): the effective volume for methods based on the SVD backbone, \(V_\text{LTXV/Wan}\): the effective volume for methods that use LTXV or Wan as a backbone, \(F_\text{SVD}\): the latent number of conditioning frames used in methods based on the SVD backbone, \(F_\text{LTXV}\): the latent number of conditioning (or input) frames used in methods based on the LTXV backbone and \(F_\text{Wan}\): the latent number of conditioning (or input) frames used in methods that use the Wan backbone.}
\label{tab:matched_volumes_compact}
\end{table}
\vspace*{1em}

We characterize the VAE latent space by \(c\times f\times h\times w\) (the number of channels, the number of frames,  height, and width):
SVD \(4\times1\times8\times8\), LTXV \(128\times8\times32\times32\), Wan \(16\times4\times8\times8\). Table~\ref{tab:matched_volumes_compact} lists representative configurations used in the main-text GPU memory study across five effective volumes, including effective volumes for SVD, LTXV, and Wan, and the latent number of conditioning (or input) frames for each run. For consistency, SVD-based methods are run at  \(576 \times 1024\) resolution, and LTXV/Wan-based methods at \(480\times832\). 

\section{Extended Experimental Results}
\label{appendix:eval_vbench}

\subsection{Additional Quantitative Comparisons}

\subsubsection{Evaluation Results Using VBench Metrics} 

This section reports additional quantitative results on OpenVid \cite{nan2024openvid} and NuScenes \cite{caesar2020nuscenesmultimodaldatasetautonomous} using the following metrics from VBench \cite{huang2023vbench}: subject consistency, background consistency, and motion smoothness (Table~\ref{tab:tab_vbench}). Table~\ref{tab:tab_vbench} also reports efficiency metrics, 
measuring total NFE and GPU memory scaling; full details are provided in the main paper and in 
Appendix~\ref{appendix:details_mem_eff}.

On OpenVid, FlowC2S (Wan) surpasses CausVid in subject consistency, background consistency, and motion smoothness, while requiring only half the input dimensionality. On NuScenes, the same trend holds for background consistency and motion smoothness, with a negligible drop in subject consistency.

\begin{table*}[t]
\centering
\begin{tabular}{l cccccc ccc}
\toprule
\multirow{3}{*}{Method}
  & \multicolumn{6}{c}{Visual Quality} & \multicolumn{3}{c}{Efficiency} \\
\cmidrule(lr){2-7}\cmidrule(lr){8-10}
  & \multicolumn{3}{c}{OpenVid} & \multicolumn{3}{c}{NuScenes}
  & \multirow{2}{*}{\shortstack{Total \\ NFE $\downarrow$}}
  & \multirow{2}{*}{$k$ (MB\,/\,$10^6$) $\downarrow$}
  & \multirow{2}{*}{$b$ (MB)} \\
\cmidrule(lr){2-4}\cmidrule(lr){5-7}
  & \shortstack{Subj. \\ Cons.$\uparrow$} & \shortstack{Bg. \\ Cons.$\uparrow$} & \shortstack{Mot. \\ Smooth.$\uparrow$}
  & \shortstack{Subj. \\ Cons.$\uparrow$} & \shortstack{Bg. \\ Cons.$\uparrow$} & \shortstack{Mot. \\ Smooth.$\uparrow$}
  & & & \\
\midrule
CausVid \cite{yin2025causvid}
  & 97.30 & 96.31 & 99.09
  & \textbf{94.25} & 94.16 & 97.85
  & 5 & 93.35 & 4003.58 \\
Ours
  & \textbf{98.37} & \textbf{98.43} & \textbf{99.29}
  & 92.89 & \textbf{95.77} & \textbf{98.25}
  & 5 & \textbf{48.64} & 3998.74 \\
\bottomrule
\end{tabular}
\caption{Quantitative comparison on OpenVid and NuScenes (val) using VBench Metrics. Subj.: Subject, Bg.: Background, Cons.: Consistency, Mot.: Motion, Smooth.: Smoothness.}
\label{tab:tab_vbench}
\end{table*}
\vspace*{2em}

\subsubsection{Comparison with Bi-Flow}

Bi-flow \cite{Liu_2025_ICCV} augments flow matching with noise perturbations at training, similar to bridge models \cite{zhou2023denoisingdiffusionbridgemodels, shi2023diffusionschrodingerbridgematching, peluchetti2023nondenoisingforwardtimediffusions, albergo2025stochasticinterpolantsunifyingframework} and learns a joint objective to predict both the velocity and noise. Bi-flow generates a video autoregressively from a single-frame, one frame at a time. Yet modeling an entire video segment by a single frame discards the inter-frame structure that models rely on to produce coherent motion, leading to temporal artifacts that compound over extended sequences. 

Table~\ref{tab:biflow_comparison} reports a quantitative comparison between Bi-Flow and FlowC2S on the OpenVid and NuScenes validation sets, with both methods trained from the LTXV backbone. Since Bi-Flow operates in a frame-by-frame autoregressive manner, generating a 17-frame continuation requires 85 total NFEs (5 NFEs per frame), whereas FlowC2S produces the entire chunk in a single pass with only 5 NFEs. Despite this substantial reduction in NFEs, FlowC2S outperforms single-frame Bi-flow across all metrics.

\begin{table*}[t]
\centering
\begin{tabular}{l ccc ccc}
\toprule
\multirow{2}{*}{Method}
  & \multicolumn{3}{c}{OpenVid} & \multicolumn{3}{c}{NuScenes} \\
\cmidrule(lr){2-4}\cmidrule(lr){5-7}
  & FID $\downarrow$ & FVD $\downarrow$ & $\mathrm{LPIPS_s}$ $\downarrow$
  & FID $\downarrow$ & FVD $\downarrow$ & $\mathrm{LPIPS_s}$ $\downarrow$ \\
\midrule
Bi-flow \cite{Liu_2025_ICCV}
  & 3.34 & 674.12 & 0.34
  & 3.99 & 873.96 & 0.53 \\
Ours
  & \textbf{0.48} & \textbf{129.20} & \textbf{0.23}
  & \textbf{1.13} & \textbf{185.48} & \textbf{0.41} \\
\bottomrule
\end{tabular}
\caption{Quantitative comparison with Bi-flow on OpenVid and NuScenes (val) using the LTXV backbone. FlowC2S is reported at 5 NFE. Lower is better ($\downarrow$); best scores in bold.}
\label{tab:biflow_comparison}
\end{table*}
\vspace*{2em}

\subsubsection{Effect of Chunk Size}
We validate the importance of multi-frame video chunks in FlowC2S by ablating the chunk size $L$ used during training, including the single-frame  case ($L=1$). All models use the LTXV backbone, are trained on OpenVid for 1406 steps, and evaluated with 17 input and 17 generated frames at inference for a controlled comparison.

Results are summarized in \cref{tab:seg_gran}. Increasing $L$ consistently improves across all metrics, demonstrating that multi-frame temporal context is essential 
for high-quality and efficient video continuation -- a single frame 
is insufficient to capture the inter-frame video structure.

\begin{table}[t]
\centering
\begin{tabular}{c ccc}
\toprule
$L$ & FID $\downarrow$ & FVD $\downarrow$ & $\mathrm{LPIPS_s}$ $\downarrow$ \\
\midrule
1  & 6.81 & 1390.84 & 0.39 \\
17 & 0.52 & 146.94  & 0.31 \\
41 & \textbf{0.50} & \textbf{126.54} & \textbf{0.24} \\
\bottomrule
\end{tabular}
\caption{Impact of chunk size $L$ at training on OpenVid. All models use the LTXV backbone with 5 NFEs. Lower is better ($\downarrow$); best scores in bold.}
\label{tab:seg_gran}
\end{table}
\vspace*{1em}

\subsection{Per-Category Analysis of TI}
\label{appendix:per_cat}

To analyze the influence of TI on generated video quality, this section reports FID and FVD across NFEs on the categorized OpenVid \cite{nan2024openvid} validation set. First, we detail the categorization procedure; then we present the full per-category results. We categorize our OpenVid validation set along two axes: motion intensity and camera motion type, yielding a 12 categories. 

\textit{Motion intensity} is determined by the motion score provided in the OpenVid metadata, which quantifies the magnitude of scene and object motion. We partition videos into the following three equally-sized groups: slow, medium, and fast (the 33rd and 67th percentiles of the motion score distribution serve as boundaries). 

\textit{Camera motion type} is also taken from the metadata of OpenVid. We define four groups based on keyword matching: (i) static—no camera movement; (ii) pan/tilt—horizontal or vertical camera rotation; (iii) zoom—focal length changes; and (iv) complex—two or more simultaneous camera motion types (e.g., pan with zoom).

\begin{figure*}[t]
    \centering
    \includegraphics[width=1.0\linewidth]{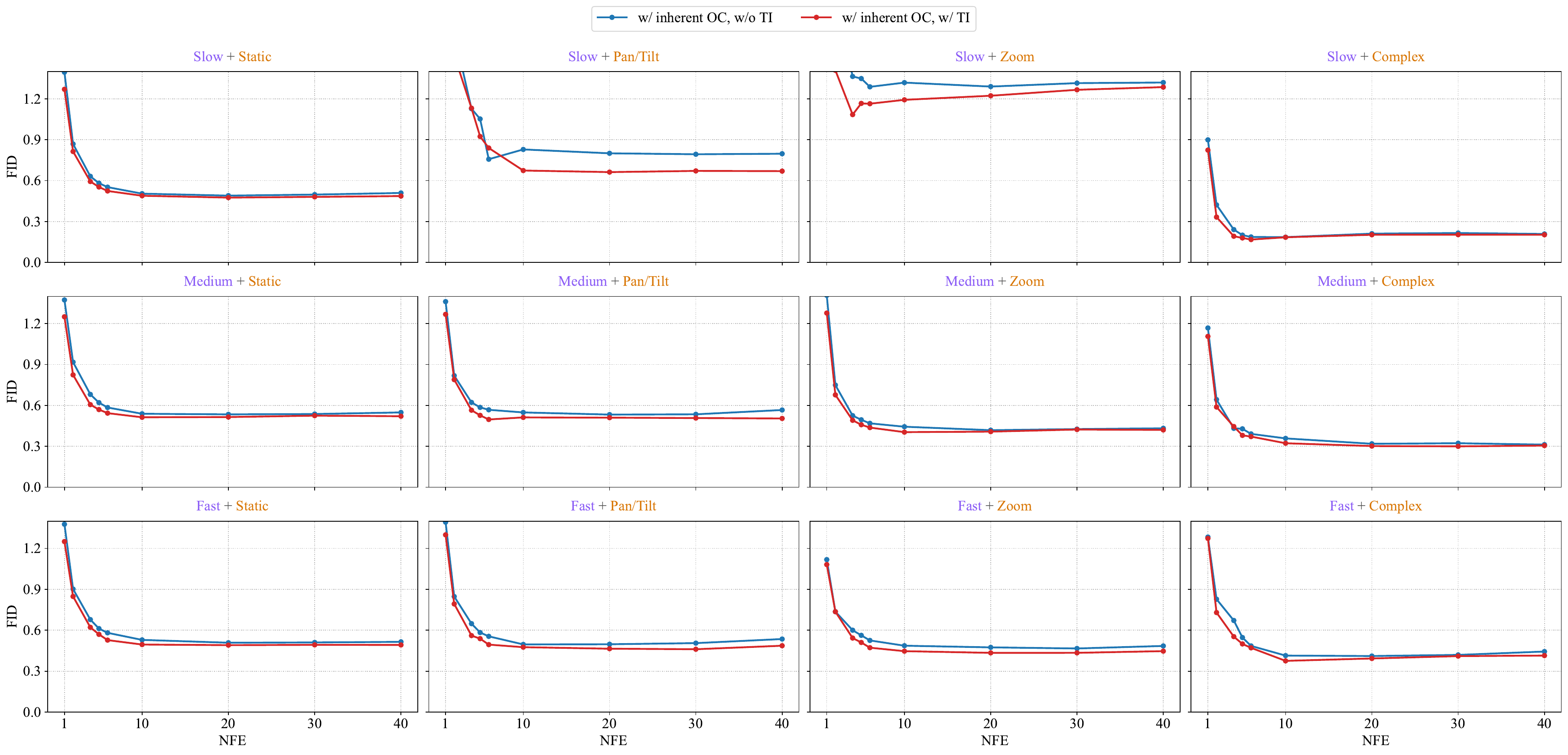}%
    \caption{Per-category FID vs.\ NFE comparing w/ inherent OC, w/o~TI (blue) and w/ inherent, OC~w/~TI (red). The benefit of TI is strongest under Pan/Tilt camera motion and Fast+Zoom and  negligible under static camera.}
    \label{fig:fig_cat_full_fid}
\end{figure*}

\begin{figure*}[t]
    \centering
    \includegraphics[width=1.0\linewidth]{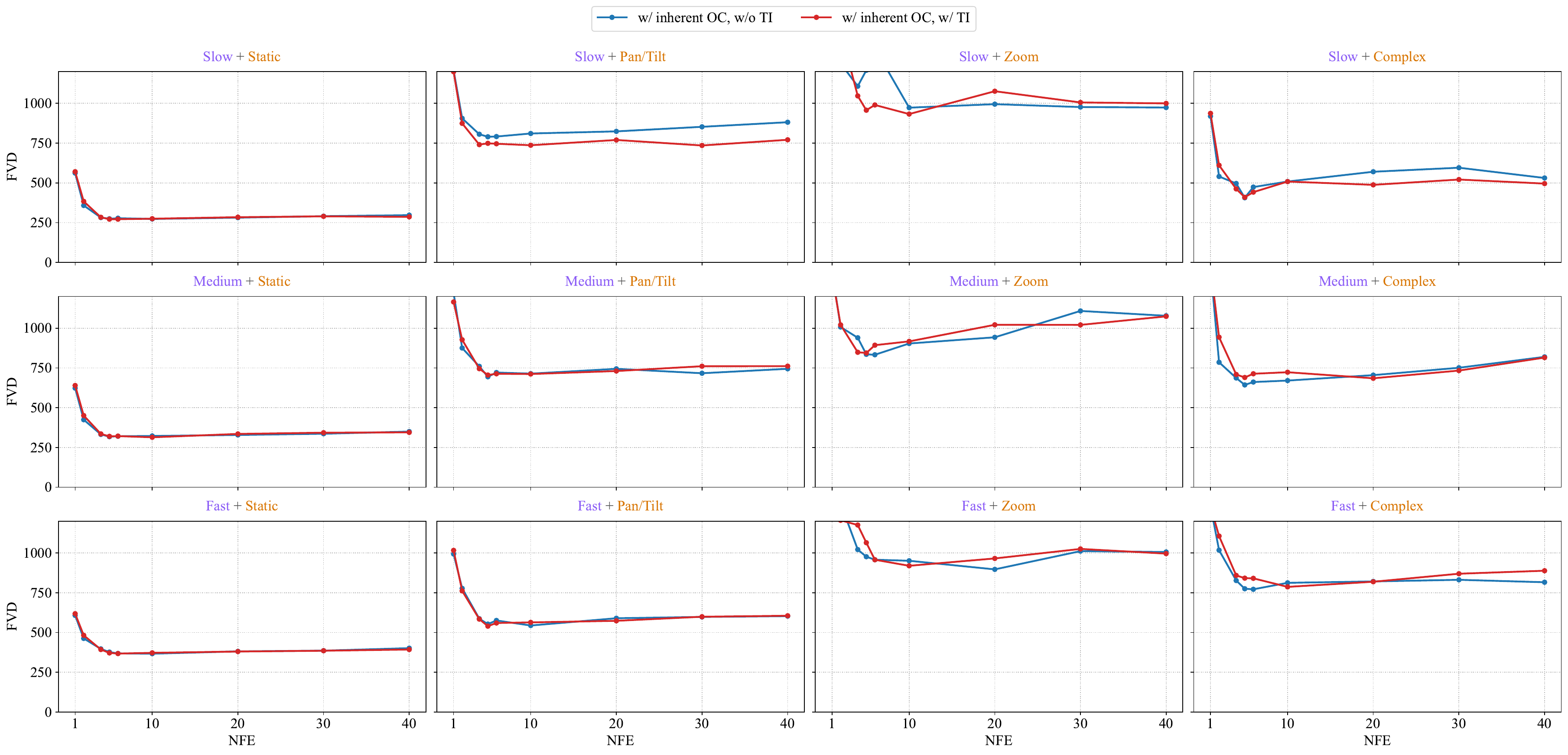}%
    \caption{Per-category FVD vs.\ NFE comparing w/ inherent OC, ~w/o~TI (blue) and w/ inherent, OC~w/~TI (red). FVD is substantially higher under zoom than other camera types. Under a static camera, the two variants are indistinguishable.}
    \label{fig:fig_cat_full_fvd}
\end{figure*}

Figures~\ref{fig:fig_cat_full_fid} and~\ref{fig:fig_cat_full_fvd} isolate the contribution of TI by comparing w/ inherent OC, ~w/o~TI (blue) against w/ inherent OC, ~w/~TI (red) across all twelve categories. We make the following observations. First, the benefit of TI is strongly conditioned on camera motion type, and FID and FVD respond to TI differently. 

Second, FID values remain consistent across most categories, with the exception of Slow~+~Zoom, which yields a higher FID than the remaining categories; in this case, w/ inherent OC, ~w/~TI achieves lower FID than w/ inherent OC, ~w/o~TI. FVD values remain consistent across static camera categories, but are higher under zoom. The following paragraphs discuss each category in detail.

\textit{Static categories.} Across all three motion speeds, w/ TI and w/o TI are nearly indistinguishable in both FID and FVD. This indicates that TI provides no measurable benefit when inter-chunk displacement is negligible. Under static camera conditions, the temporal coupling alone provides a sufficiently well-conditioned source distribution, and the vector field near the timestep zero requires no additional calibration.

\textit{Pan/Tilt categories}.TI helps most in this category, on both metrics. In FID, red lies below blue across all speeds, with the gap most pronounced in Slow~+~Pan/Tilt. In FVD, the same ordering holds: red is below blue and remains stable across NFE, while blue shows a mild upward drift at higher NFE. Pan/Tilt motion produces globally translated but otherwise predictable target latents — the regime where we hypothesize that structured camera motion produces more predictable target latents, making the inverted initialization more informative than in other categories.

\vspace*{2em}

\vspace*{2em}

\textit{Zoom categories.} In FID, red is generally below blue — the benefit of TI is present but moderate, and in Slow~+~Zoom the two curves actually cross, with blue outperforming red at intermediate NFE before red recovers at higher NFE. In FVD, w/o TI degrades with increasing NFE in Medium~+~Zoom and Fast~+~Zoom, while w/ TI remains comparatively stable or improves slightly. 

\textit{Complex categories.} In FID, w/ TI is at or below blue across all speeds, with the gap widening for medium and fast speeds. In FVD, w/ TI and w/o TI trade positions across NFE in several panels, most visibly in Fast~+~Complex, where neither curve dominates cleanly. 

The complex category represents a mix of multiple camera motion types. We hypothesize that the predictability assumption underlying TI, that structured motion constrains the source distribution, is weakest here, since complex motion produces less predictable inter-segment transitions. The partial FVD benefit nevertheless suggests TI still improves quality on average, even if the temporal coherence benefit is less stable than under structured camera types.

\section{Ablation Algorithms}
\label{appendix:appendix_more_algos}

In addition to the main training algorithm described in the paper, in this section, we include two auxiliary variants used in our ablation studies for the assessment of the primary design decisions of FlowC2S. Specifically, we provide pseudocode for: fine-tuning without Inherent Optimal Couplings (OC) and without Target Inversion (TI) (Algorithm~\ref{algo:algo-2}); and fine-tuning with Inherent OC but without Target Inversion (Algorithm~\ref{algo:algo-3}).

In Algorithm~\ref{algo:algo-2}, current and succeeding frames are sampled independently from their respective distributions, with no target inversion applied. By contrast, Algorithm~\ref{algo:algo-3} incorporates Inherent OC, ensuring coupling between current and succeeding frame chunks as described in the main text, while still omitting target inversion.

\section{Additional Visual Results}
\label{appendix:more_vis_results}

\begin{figure*}[t]
    \centering
    \includegraphics[width=1.0\linewidth]{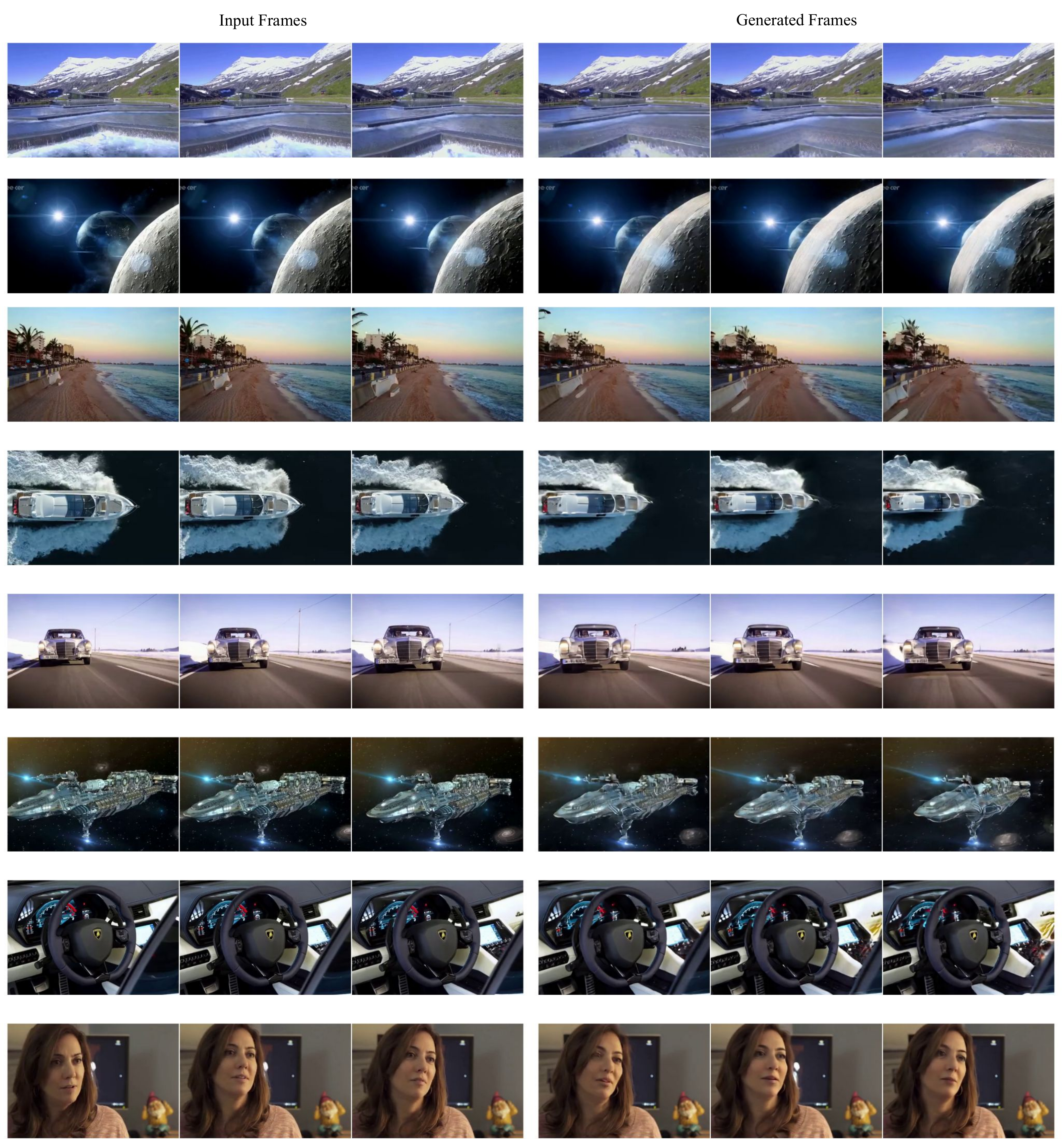}%
    \caption{Additional visual results on OpenVid (val). FlowC2S, fine-tuned from LTXV, generates video continuations that are both temporally coherent and visually plausible, maintaining consistency with the content of the input frames. Frames shown at stride 20. 
    }
    \label{fig:fig_8}
\end{figure*}

\begin{figure*}[t]
    \centering
    \includegraphics[width=1.0\linewidth]{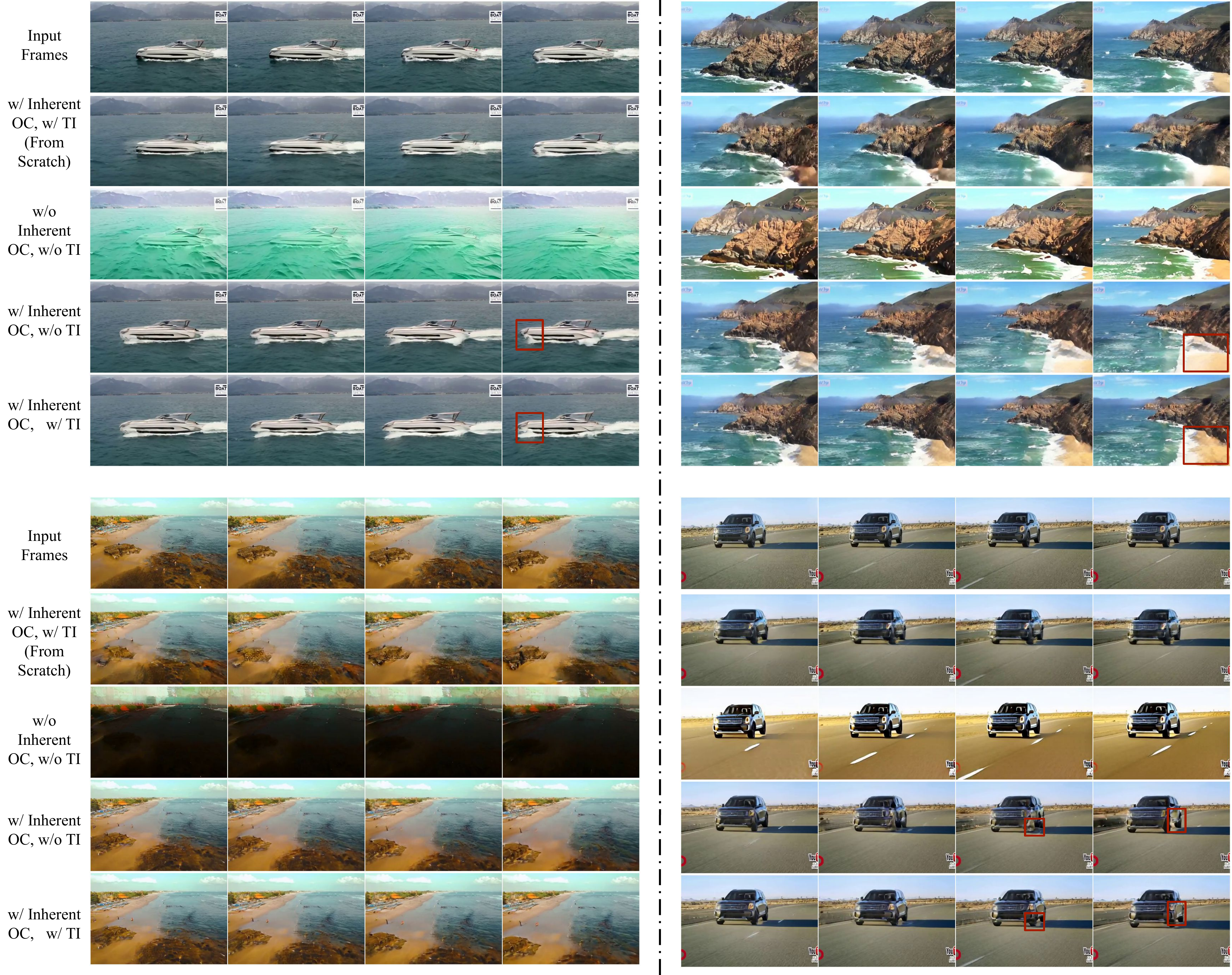}%
    \caption{Additional visual results on ablation across four training setups (frames shown with stride 13). Training from scratch with OC+TI introduces visual artifacts, manifested as noticeable interpolation effects between the given input frames and the generated outputs. Fine-tuning without OC+TI results in generated videos with temporal inconsistencies with the current frames and color shifts. Incorporating both OC and TI during fine-tuning yields sharper details than OC-only fine-tuning (see red boxes). 
    }
    \label{fig:fig_9}
\end{figure*}

\begin{figure*}[t]
    \centering
    \includegraphics[width=1.0\linewidth]{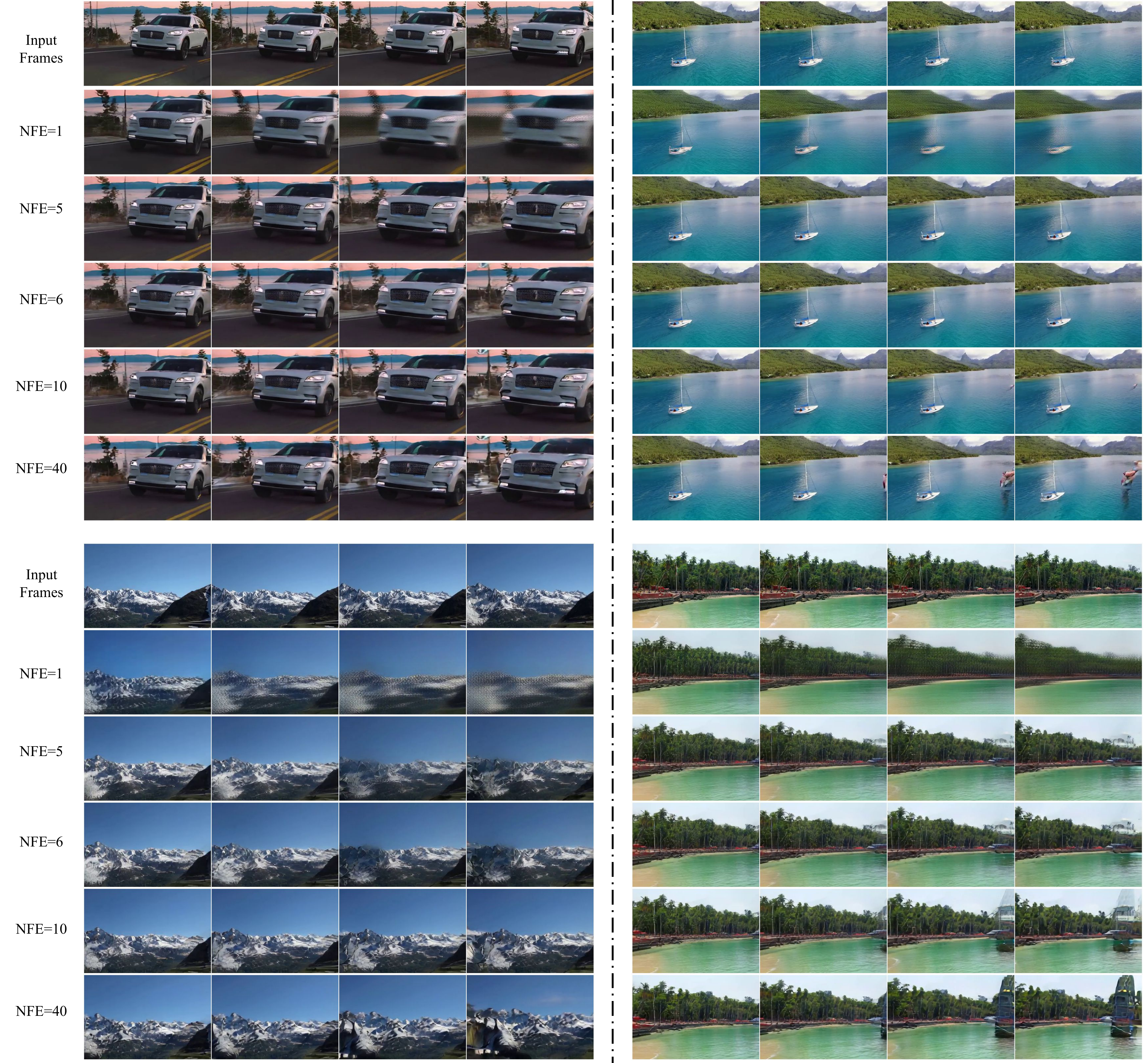}%
    \caption{Ablation on neural function evaluations (NFEs). Frames are shown with a stride of 13. 5–10 NFEs yield quality comparable to 40 NFEs, whereas a single NFE produces blurry, degraded outputs.  
    }
    \label{fig:fig_10}
\end{figure*}

This section presents additional qualitative results of generated  continuations from FlowC2S (Fig.~\ref{fig:fig_8}), visual comparisons from ablations on its principal design choices (Fig.~\ref{fig:fig_9}), as well as studies varying the number of neural function evaluations (NFEs) (Fig.~\ref{fig:fig_10}) and the number of frames in the generated videos (Fig.~\ref{fig:fig_11}). The current video chunks are taken from the OpenVid validation set, and the continuations are generated with our model, fine-tuned from LTXV.  We set the number of input and generated frames to \(41\) and resolution to \(256 \times 384\) in the examples provided in Fig.~\ref{fig:fig_8}, Fig.~\ref{fig:fig_9}, and Fig.~\ref{fig:fig_10}.

Fig.~\ref{fig:fig_8} shows the input frames and the generated continuations by our method. FlowC2S produces video continuations that exhibit detailed and strong temporal coherence with the observed context; all achieved without explicit conditioning mechanisms, but simply by taking the current frames directly as an input to the model. 

For instance, the motion of the boat traversing left to right with accompanying water dynamics is reproduced with coherent detail. In another case, the car advancing from the background toward the foreground continues seamlessly in the generated continuation, maintaining consistency with both camera dynamics and temporal flow. Thus,  FlowC2S preserves logical structure and temporal realism in generated videos.

Fig.~\ref{fig:fig_9} presents the current frames as input and the results produced by the model trained by the following four configuration variants: training from scratch with Inherent OC and TI, fine-tuning from LTXV without OC and TI, fine-tuning with OC but without TI, and fine-tuning with both OC and TI. 

Training from scratch with OC and TI leads to blurring and visual artifacts and close similarity with the provided input frames, reflecting poor convergence and the significance of using a pre-trained model as an initialization. Similarly, fine-tuning without OC and TI produces unstable results, with noticeable artifacts and color discrepancies (e.g., in the boat and car examples). By contrast, incorporating both OC and TI during fine-tuning yields markedly improved visual fidelity. For instance, in the sea-and-mountain example, the inherent OC-only setting generates blurred sand textures, while the joint inherent OC and TI setting restores sharper, more realistic details, as highlighted by red boxes in Fig.~\ref{fig:fig_9}.

\begin{algorithm}[t]
\caption{Flowing From Current To Succeeding Frames (w/o Inherent OC, w/o Target Inversion)}
\begin{algorithmic}[1]
\State \textbf{Require:} \text{pretrained} $u_t^{\theta^*}$
\State $u_t^{\theta} \gets u_t^{\theta^*}$
\For{$x_0 \sim p_{\text{c\_fc}}, x_1 \sim p_{\text{s\_fc}} $}
    \State $\mu_1, \sigma_1 = \text{VAE}(x_1), \quad x_1 \gets \mu_1$
    \State $\mu_0, \sigma_0 = \text{VAE}(x_0)$, \quad $x_0 \gets \mu_0$
    \State $t \sim \mathcal{U}[0, 1]$
    \State $x \gets (1 - t)x_0 + tx_1$
    \State $\mathcal{L}_{\text{CFM}}(\theta) = \left\| u_t^\theta(x) - (x_1 - x_0) \right\|^2$
    \State Update $\theta$ using GD on $\mathcal{L}_{\text{CFM}}(\theta)$
\EndFor
\end{algorithmic}
\label{algo:algo-2}
\end{algorithm}

Fig.~\ref{fig:fig_10} illustrates the input frames and the videos generated by the proposed FlowC2S method with 1, 5, 6, 10, and 40 NFEs. As shown, a single NFE is inadequate for generating video continuations, yielding results with severe motion blur (e.g., in the white car sequence). In contrast, using 5–10 NFEs produces video continuations of competitive fidelity to that obtained with 40 NFEs, demonstrating that our method achieves good visual quality with reduced NFEs.

Fig.~\ref{fig:fig_11} depicts the input frames and generated video results of generating long video continuations. In this example, the number of current and generated frames by the model is set to 113. Although FlowC2S is trained only on 17- and 41-frame video chunks, it generalizes to substantially longer video continuations. For example, in the tractor sequence, its generated motion from backward to forward constitutes a plausible temporal continuation with respect to the provided input frames.

\vspace*{1em}

\section{Limitations and Future Work}
\label{appendix:lim_and_future_work}

FlowC2S delivers a \(2\times\) reduction in input dimensionality and achieves state-of-the-art video continuation in quantitative metrics with substantially fewer NFEs, yet important limitations remain. Below, we elaborate on the constraints outlined in the main text and describe corresponding avenues for future work.

\vspace*{1em}

\begin{algorithm}[t]
\caption{Flowing From Current To Succeeding Frames (w/ Inherent OC, w/o Target Inversion)}
\begin{algorithmic}[1]
\State \textbf{Require:} \text{pretrained} $u_t^{\theta^*}$, \(\Pi ( p_{\text{c\_fc}}, p_{\text{s\_fc}})\)
\State $u_t^{\theta} \gets u_t^{\theta^*}$
\For{$x_0, x_1 \sim \Pi$ \textbf{with} $(x_0, x_1)$ \textbf{inherent optimal couplings}}
    \State $\mu_1, \sigma_1 = \text{VAE}(x_1), \quad x_1 \gets \mu_1$
    \State $\mu_0, \sigma_0 = \text{VAE}(x_0)$, \quad  $x_0 \gets \mu_0$
    \State $t \sim \mathcal{U}[0, 1]$
    \State $x \gets (1 - t)x_0 + tx_1$
    \State $\mathcal{L}_{\text{CFM}}(\theta) = \left\| u_t^\theta(x) - (x_1 - x_0) \right\|^2$
    \State Update $\theta$ using GD on $\mathcal{L}_{\text{CFM}}(\theta)$
\EndFor
\end{algorithmic}
\label{algo:algo-3}
\end{algorithm}

\textbf{Complex Scenes and Motions.} While quantitative metrics (FID and FVD) and the visual results presented in the main text indicate that our method generates plausible video continuations for a diverse set of input frames (including natural, cinematic, and human-centric scenes), the model struggles to handle highly complex motions and abrupt scene transitions. This limitation primarily arises from our dataset construction pipeline: to mitigate abrupt scene changes, we employ a histogram-based scene change detector in order to segment videos into temporally coherent chunks, thereby simplifying the training distribution and enabling evaluation under restricted computational and data budgets. 

Therefore, future work will explore training the proposed method on less constrained datasets that retain complex scenes with abrupt transitions, with the goal of improving the model's generalization and robustness to challenging motion and scene dynamics.

\begin{figure*}[t]
    \centering
    \includegraphics[width=1.0\linewidth]{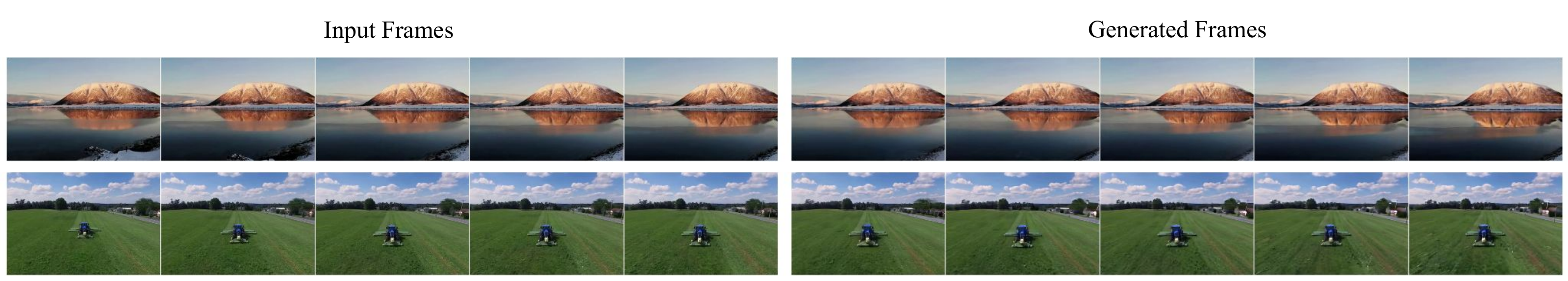}%
    \caption{Long video continuation. The number of input and future frames is 113, and the frames are visualized with a stride of 28. Despite being trained only on 17- and 41-frame sequences, FlowC2S successfully generates coherent long-video continuations. 
    }
    \label{fig:fig_11}
\end{figure*}

\begin{figure*}[t]
    \centering
    \includegraphics[width=1.0\linewidth]{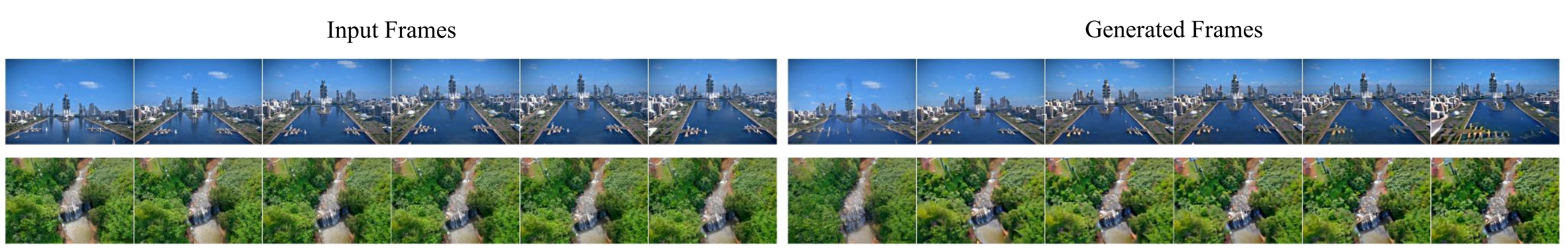}%
    \caption{Failure cases for very long continuation. Shown are 129 input and generated frames (visualized with a stride of 28). Beyond ~113-frame video chunks, FlowC2S degrades, exhibiting interpolation artifacts and reduced coherence with the given context.
    }
    \label{fig:fig_12}
\end{figure*}

\textbf{Heterogeneous Video Chunk Lengths.} In the main text, we showed that FlowC2S maintains stable FID and FVD and strong visual quality for long video continuation with visuals in Fig.~\ref{fig:fig_11} with 113 current and generated frames; despite training solely on 17- and 41-frame chunks. Beyond this range, however, results become degraded with visible interpolation artifacts with the input frames. Fig.~\ref{fig:fig_12} illustrates representative failure cases at longer video continuation (129 input and generated frames).

Consistent with prior evidence in text-to-image \cite{podell2023sdxlimprovinglatentdiffusion, esser2024scalingrectifiedflowtransformers, dai2023emuenhancingimagegeneration, xie2024sanaefficienthighresolutionimage} and text-to-video \cite{zheng2024opensorademocratizingefficientvideo, peng2025opensora20trainingcommerciallevel, yang2025cogvideoxtexttovideodiffusionmodels, Wan2025Wanopenadvancedlargescale, kong2025hunyuanvideosystematicframeworklarge, hacohen2024ltxvideorealtimevideolatent, zhou2024allegroopenblackbox, chen2025gokuflowbasedvideo, seawead2025seaweed7bcosteffectivetrainingvideo} literature, exposure to diverse spatial and temporal scales during training improves inference-time robustness of diffusion and flow-based models. Thus, a natural next step is therefore to train the proposed FlowC2S approach with heterogeneous video chunk lengths, broadening temporal coverage of the model and strengthening generalization for the generation of very long video continuations.

\textbf{Controllability.} While the proposed method introduces a new perspective on video continuations, flowing directly from current to succeeding frames, yielding both a twofold reduction in input dimensionality and fewer neural function evaluations compared to existing methods, the current framework does not leverage additional conditioning signals such as text prompts, depth maps, motion or camera trajectories.

After the arrival of diffusion \cite{pmlr-v37-sohl-dickstein15, song2021denoising, NEURIPS2020_4c5bcfec, song2021scorebased, NEURIPS2021_49ad23d1} and flow \cite{lipman2023flowmatchinggenerativemodeling, liu2022flowstraightfastlearning, albergo2023buildingnormalizingflowsstochastic} models, controllability has become a key in image and video generation \cite{
zhang2023addingconditionalcontroltexttoimage, 
zhang2023controlvideotrainingfreecontrollabletexttovideo,
jiang2025vaceallinonevideocreation, he2025cameractrlenablingcameracontrol, 
Lei_2025_CVPR, xiao2024omnigenunifiedimagegeneration, xu2024camcocameracontrollable3dconsistentimagetovideo, liang2024movideomotionawarevideogeneration, zheng2025vidcraft3cameraobjectlighting, hou2025trainingfreecameracontrolvideo, ma2025controllablevideogenerationsurvey, zhou2025lightavideotrainingfreevideorelighting}. Extending FlowC2S to incorporate auxiliary signals is therefore a promising direction for future work, enabling richer, more adaptable, and user-guided generation of video continuations; in particular, augmenting the current architecture with conditioning signals including scene geometry (e.g., depth or normal maps), object-level motion cues, and camera trajectories could provide fine-grained spatial and temporal control over the synthesized content and better align the generated video continuations with user intent and downstream application constraints.



\end{document}